# Vision Transformers in Medical Imaging: A Review

Emerald U. Henry[a*], Onyeka Emebo[b,c], Conrad Asotie Omonhinmin[d]

[a]*College of Engineering, Department of Mechanical Engineering, Covenant University, Ota, Ogun State, Nigeria.*
[b]*College of Engineering, Department of Computer Science, Virginia Tech, Blacksburg Virginia, USA*
[c]*College of Science and Technology, Department of Computer & Information Science, Covenant University, Ota, Ogun State, Nigeria*
[d]*College of Science and Technology, Department of Biological Sciences/Biotechnology cluster, Covenant University, Ota, Ogun State, Nigeria*

**ABSTRACT**

Transformer, a model comprising attention-based encoder-decoder architecture, have gained prevalence in the field of natural language processing (NLP) and recently influenced the computer vision (CV) space. The similarities between computer vision and medical imaging, reviewed the question among researchers if the impact of transformers on computer vision be translated to medical imaging? In this paper, we attempt to provide a comprehensive and recent review on the application of transformers in medical imaging by; describing the transformer model comparing it with a diversity of convolutional neural networks (CNNs), detailing the transformer based approaches for medical image classification, segmentation, registration and reconstruction with a focus on the image modality, comparing the performance of state-of-the-art transformer architectures to best performing CNNs on standard medical datasets.

## 1 Introduction

The transformer (Vaswani et al., 2017), identified by its attention mechanism, has become the dominant deep learning architecture in the field of natural language processing (NLP) due to its success in text to speech translation (Li et al., 2019), natural language generation (Topal et al., 2021), text synthesis ( Li et al., 2019) and speech recognition (Feng et al., 2022). The success of transformers in the field of natural language processing owes primarily to their ability to capture long range dependencies that aids in the retention of contextual information (Vaswani et al., 2017) as opposed to recurrent neural networks (RNNs) (Graves et al., 2013; Sak et al., 2014) that utilizes a sequential inference process and cannot efficiently capture long range dependencies. A plethora of transformer architectures for natural language processing have been proposed since 2017, a few of the popular architectures are; Bidirectional Encoder Representation from Transformer (BERT) and its variants (Devlin et al., 2019; Lan et al., 2019; Liu et al., 2019), Generative Pre-Trained Transformer (GPT) and its variants (Hoppe & Toussaint, 2020; Radford et al., 2018; Winata et al., 2021).

In the computer vision field, convolutional neural networks (CNNs) have achieved efficient performance mainly due to the structure of their architectures (He et al., 2016; Louis, 2013; Peng et al., 2021; Yang et al., 2022; Y. Zhang et al., 2020). Recent evidence show CNNs exploits the locality of pixels aiding capture of vision semantics and yield acceptable performance even on small datasets (d'Ascoli et al., 2021), CNNs are also known to possess progressively enlarge receptive field that aids in the representation of image hierarchical structure in form of semantics. However, the advent of transformers apprised researchers of CNNs' major drawback, the inability to capture long range dependencies such as the extraction of contextual information and the non-local correlation of objects (Zhang et al., 2020). This has led to attempts to incorporate self-attention either spatially (Cao et al., 2019; Huang et al., 2019; Xiaolong Wang et al., 2018) or channel-wise (Hu et al., 2020; Q. Wang et

*Corresponding Author*: Tel: +234-811-940-0230
Email Address: emerald.henry@stu.cu.edu.ng

al., 2020; Woo et al., 2018), into the conventional CNN architecture. Eventually, the first pure transformer for computer vision application named Vision Transformer (ViT) was proposed by (Dosovitskiy et al., 2020), in which they demonstrated the equivalence between multi-head self-attention attached to a multi-layer perceptron and CNNs by considering image classification as a sequence prediction task hence utilizing patch down-sampling and quadratic positional encoding in order to capture long-range dependencies between image tokens (patches). In recent literatures, these pure or hybrid vision transformers (ViTs) have achieved state-of-the-art performance over the CNN benchmarks. This has been consistent across a is variety of computer vision tasks, such as image classification (Dosovitskiy et al., 2020), image reconstruction (Jiang et al., 2021), pixel segmentation ( Zheng et al., 2021), image captioning (Cheng et al., 2021), three-dimensional imaging (Zhou et al., 2021) and video applications (Zhou et al., 2018).

CNNs have substantially influenced the field of medical imaging because of the pertinent need for classification, segmentation and detection, required of the variety of imaging modalities including ultrasound, X-ray radiography, magnetic resonance imaging (MRI), computed tomography (CT), whole-slide-images (WSIs) etc. (Darby et al., 2012). Surprisingly, about 90% of all healthcare data are compiled instances of the various medical imaging modalities. This implies that there is an outlay of data available to foster efficient modelling for clinical diagnosis and decision-making. CNNs (He et al., 2016; Hu et al., 2020; Weng & Zhu, 2021) have excelled because of the ability to learn spatio-temporal dependencies within an image and utilize this in the extraction of distinguishable representation (Li et al., 2020; Susanti et al., 2017; Yu & Helwig, 2022). However, convolutional layers have stationary weights that do not adapt for a specific input image, offer their models a limited effective receptive field that limits the ability to capture long-range dependencies between pixels.

The successes of transformers on natural images, have encouraged researchers to query further the application of self-attention in medical imaging in order to effect long-range dependencies between pixels. Additionally, transformers have achieved comparable performance to state-of-the-art CNNs on medical image classification (Xie, Zhang, Xia, et al., 2021), detection (Ghaderzadeh & Asadi, 2021), segmentation (Tragakis et al., 2022) and reconstruction (Zhou et al., 2022). Literatures have recorded transformers of better performance than state-of-the-art CNNs however the performance of transformers over CNNs is still debatable and newer modifications to the transformer architecture emerge every day in an attempt to mitigate transformer related problems.

This paper attempts to provide a detailed review on the application of transformers in medical imaging and to compare their performance with state-of-the-art CNNs.

The paper provides detailed and recent review on of transformer state-of-the-art in medical imaging and detailed comparison between CNN and transformer benchmarks. To aid in visual identification and comprehension we have included a taxonomy articulating the disparate application of transformers in medical imaging with references.

Our paper is outlined as follows: the preliminaries of the original transformer, and CNNs by detailing the various combinations with transformers as can be found in the literature. The current progress of the transformer state-of-the-art in medical image classification, segmentation, registration, and reconstruction. In addition, the paper, identifies all known positives and negatives of the transformer, and provide a detailed comparison between present state-of-the-art transformer approaches and that of CNNs.

## 2 TRANSFORMERS

### 2.1 Attention in Transformers

The fundamental transformer architecture as proposed by Vaswani *et al.* (2017) is a sequence-to-sequence model that is composed of self-attention and point-wise feed-forward network (FFN) block that extracts global dependencies between tokens

*Corresponding Author*: Tel; +234-811-940-0230
Email Address: emerald.henry@stu.cu.edu.ng

(words). The complete architecture includes layer normalization after each attention block, a linear transformation function and a softmax function. This architecture constitutes the basis for more complex and efficient, supervised or self-supervised models today, a part of this success can be attributed to the concept of *Attention*.

Attention mechanism is the primary way humans sort relevant from irrelevant data by unintentionally paying attention to some part of data set, while discarding other parts. Few scientists have attempted to build neural networks that model this behavior, initially for use in language processing tasks (Bahdanau et al., 2015; Dai et al., 2017; Xu et al., 2015). A typical attention, regarded as the "Bahdanau attention" computes a weighted sum of each feature while highlighting the most relevant features from a *feature matrix*.

*Self-attention* was designed to emphasize relationships between data regardless of their position in the sequence. It is mathematically expressed by a map function of queries, keys and values such that for each input $X \in \mathbb{R}^c, i = 1, \dots, n$ there exist a query $Q \in \mathbb{R}^{n \times d}$, a key $K \in \mathbb{R}^{n \times d}$, and a value $V \in \mathbb{R}^{n \times d}$, which are utilized in generating learning parameters $W^q, W^k, W^v$ respectively.

$$\begin{aligned} Q &= X \times W^q, & W^q &\in \mathbb{R}^{c \times d}, \\ K &= X \times W^k, & W^q &\in \mathbb{R}^{c \times d}, \quad (1) \\ V &= X \times W^v, & W^q &\in \mathbb{R}^{c \times d}, \end{aligned}$$

The output is a probability that requires normalization, usually achieved by a softmax function to attain an output distribution represented by the equation (2).

$$Attention\ (Q, K, V) = softmax\left(\frac{QK^T}{\sqrt{D_K}}\right) V = AV \quad (2)$$

The output of a self-attention block is the sum of element V and the matrix A equipping this attention block with the ability to capture global dependencies within a data.

*Multi-head self-attention* can be applied to better capture hierarchical features. These are computed in parallel after the final output is obtained by concatenating each individual attention block. This operation is similar to the use of multiple kernels within convolution operations, mathematically expressed by:

$$Z_i = Attention(Q \times W_i^q, K \times W_i^q, V \times W_i^v),$$
(3)
$$MSA(Q, K, V) = concat[Z_1, \dots, Z_h] \times W^o$$

Here $h$ represents the total number of heads and $W^o$ represents an output matrix of the concatenated projection of all self-attention $W_i^q$, $W_i^q$, $W_i^v$ of the $i^{th}$ attention.

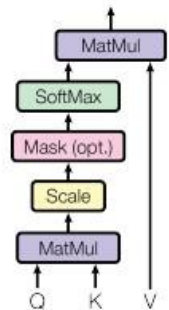
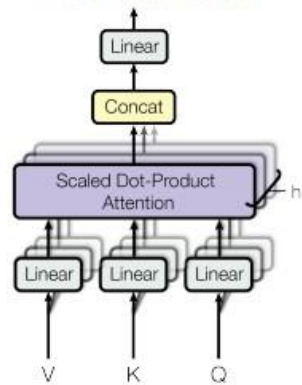

Figure 1: Multiple attention layers arranged in parallel (left) Self-Attention. (Right) Multi-Head Self-Attention (Vaswani et al., 2017).

*Corresponding Author*: Tel; +234-811-940-0230
Email Address: emerald.henry@stu.cu.edu.ng

## 2.2 Point-Wise Feed-Forward Network

The output from the multi-head self-attention block is fed into a feed-forward network comprising two linear activation function and a rectified linear unit (RELU) activation, as expressed in equation (4).

$$FFN(X) = ReLU(XW_a + B_a)W_b + B_b \quad (4)$$

Here $X$ represents the output from the previous layer and $W_a, W_b, B_a, B_b$ are trainable parameters of dimensions $D^c$ and $D^n$, represented as $W_i \in \mathbb{R}^{D^c \times D^n}$ and $B_i \in \mathbb{R}^{D^c \times D^n}$ where $i = a, b$. It is to be noted that $n$ should always be larger than $c$.

## 2.3 Positional Encoding

Learnable parameters are typically employed to aid the network retain positional information, this could be achieved by recurrence or convolutions however, in the transformer architecture this is achieved by inputting information about the position of tokens (words or patches) into the sequence. Vaswani *et al.* (2017) experimentally utilized sine and cosine functions of varying frequencies

$$PE_{(pos,i)} = \begin{cases} \sin(pos \times w_n) & if \ i = 2n \\ \cos(pos \times w_n) & if \ i = 2n+1 \end{cases} \quad (5)$$

$$w_n = \frac{1}{10000^{2n/k}}, \qquad n = 1, \ldots, k/2$$

Here $pos$ represents the initial position of the vector while $n$ represents the length of the vector, $i$ represents the particular instance. In the first pure vision transformer by (Dosovitskiy et al., 2020) the learned position is outputted as a vector and serves as input into the encoder. This vector is a sequence of n-dimension patches where $n = 1, \ldots, k$. While a lower n-value will store better positional information; it will also be computationally expensive.

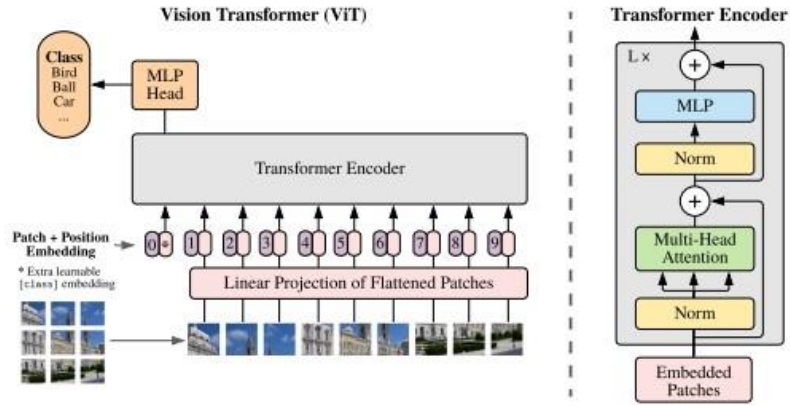

Figure 2: Vision Transformer model overview: Based on original transformer architecture (Dosovitskiy et al., 2020).

## 2.4 Vision Transformers

A variety of transformer based vision models exists, a few of the prominent ones are; Vision Transformer (ViT) (Dosovitskiy et al., 2020), Detection Transformer (DETR) (Carion et al., 2020), data-efficient image transformer (DeiT) (Touvron et al., 2020) and Swin-Transformer (Ze Liu et al., 2021). The ViT was the first pure adaptation of the vanilla transformer in the field of computer vision. It entails an encoder and a task specific decoder structure where input images are broken into a sequence of non-overlapping patches of size $(C \times P \times P)$. Here C represents the number of channel of the image and P represents its length and width. The information about the position of patches in the image is converted into a vector. This positional information, along with the sequence of non-

*Corresponding Author*: Tel; +234-811-940-0230
Email Address: emerald.henry@stu.cu.edu.ng

overlapping patches are fed into the encoder block of the transformer containing multi-head self-attention, layer normalization and a multi-layer perceptron (FFN), this is depicted in Figure (2).

DETR is a hybrid that attaches a transformer encoder to a CNN architecture; it was the first attempt at partially conveying attention from the vanilla transformer to a vision task of object detection. The swin-transformer was designed to reduce the computational cost of the ViT and was tested majorly on segmentation tasks. Liu et al., (2021) attempted varying patch sizes so as to reduce the requirement of full attention due to the input sequence of non-overlapping patches; they introduced the shifted window attention that creates patches of various sizes in a hierarchical order, and further assists in preserving spatial information.

The efficiency of the ViT could only be observed during large scale training as it performs poorly on small datasets, and Touvron et al., (2020) proposed the DeiT in an attempt at solving this problem. DeiT adopts a CNNs teacher and a transformer student structure in a knowledge distillation framework in which a distillation token is added for the purpose of learning from the teacher model. This knowledge is then inherited by the student model while imparting inductive bias.

## 2.5  Hybrids

Hybrids including DETR and DeiT can be grouped into three categories, according to the work of Jun Li et al., (2022). These categories are; *ConvNet-like-Transformer*, *Transformer-like-ConvNets* and *Transformer-ConvNet hybrids.*

*ConvNet-like-Transformers* are vision transformers that inherit the properties of conventional CNNs with the aim of improving the efficiency of the transformer. The DeiT (Touvron et al., 2020) is an example of this type of transformer, it attempts to develop transformers that inherit the inductive bias present in convolutional neural networks. Other examples include; Swin Transformer (Ze Liu et al., 2021), HaloNets (Vaswani et al., 2021), DAT (Xia et al., 2022) and PVT (Wang et al., 2021). *Transformer-like-ConvNets* are convolutional neural networks that inherit some of the properties of transformers by sparing or partial integration of transformers into the architecture. The most prominent properties researchers try to integrate to CNNs is self-attention from transformers. Few example include; CoT (Yehao Li et al., 2022), BoTNet (Srinivas et al., 2021) and ConvNext (Zhuang Liu et al., 2022). *Transformer-ConvNet hybrids* try to form architectures that consist of convolutions, multi-layer perceptrons and multi-head self-attention blocks in an attempt to fully leverage the strength of both architectures and form models that are more efficient. A few examples include CvT (Wu et al., 2021), Mobile-former (Chen et al., 2021), Conformer (Peng et al., 2021), CoAtNet (Dai et al., 2021) and ConViT (d'Ascoli et al., 2021).

## 3  TRANSFORMER IN MEDICAL IMAGING

### 3.1  Datasets

Transformers are generally known to perform better in large training than small sized training due to the absence of inductive bias that bolsters few shot learning. In an attempt to solve this problem, researchers have proposed several hybrid architectures that seeks to incorporate strengths of the convolutional neural network into the transformer. The availability of public medical datasets of diverse modalities have been a major deterrent to the training and re-training of state-of-the-art CNN architectures like ResNet (K. He et al., 2016) and EfficientNet (Tan & Le, 2019) for the development of domain specific weights to serve as a feature extractor layer in transfer learning for both CNNs and Transformers. A few researchers have proven that transformers benefit more from transfer learning than CNNs (Caron et al., 2021; Raghu et al., 2019, 2021). However, standard weights like ImageNet do not serve as efficient feature extractors for medical imaging tasks across architectures (Hosseinzadeh Taher et al., 2021) hence the need for a detailed outline of publicly available medical datasets. The detailed compilation

*Corresponding Author*: Tel; +234-811-940-0230
Email Address: emerald.henry@stu.cu.edu.ng

of most of the publicly available datasets of various medical image modalities with their description; their download link is provided in within their publication in Table 1 (Parvaiz et al., 2022).

Table 1: Compilation of published available dataset of medical image modalities

| | | DETECTION |
|---|---|---|
| **Modality** | **Dataset** | **Description** |
| **Histopathology Images** | Cancer Genome Atlas (Weinstein et al., 2013) | The dataset represents heat maps of 33 tumor types, and 3 distinct expressions; reverse-phase protein array, gene expression and miRNA. |
| **CT-scans** | COVID-19 CT-2A (Gunraj et al., 2022) | A benchmark dataset containing 3 classes, COV19 pneumonia, non-COV19 pneumonia and normal. Comprises data collated from 15 countries and contains about 4,500 samples. |
| | COVID-19 CT-DB (Kollias et al., 2021) | Contains annotated data that indicates the existence of COVID-19 |
| **X-rays** | COVIDx (Gunraj et al., 2022) | A dataset of 3 classes: COVID positive, viral pneumonia and normal images. |
| | COVIDGR-E (Tabik et al., 2020) | Contains about 430 images of COVID-19 pneumonia |
| **Fundus Images** | IDRiD (Saeed et al., 2021) | A dataset comprising 81 images for Micro-aneurysm detection. |

| | | CLASSIFICATION |
|---|---|---|
| **Modality** | **Dataset** | **Description** |
| **Histopathological Images** | Cancer Histology Dataset (Kather et al., 2016) | These is a dataset comprising whole-slide-images of colorectal cancer cells. |
| **CT-scan** | COVID-CT-set (Rahimzadeh et al., 2020) | A database contain a large number of lung scans for covid-19 classification. |
| | Sars-CoV-2 (Soares & Angelov, 2020) | A dataset containing multiple lung scans for classifying COVID-19 SARS-variant. |
| | COVID-19-CTDB (Kollias et al., 2021) | An annotated dataset of chest scans. |
| | COVID-CT (X. Yang et al., 2020) | A dataset containing a number of chest scans complied from a number of literature. |
| | CT-emphysema-DB (Sørensen et al., 2010) | A dataset of 115 high-resolution scans and 168 manually annotated square patches. |
| | LUNA16 (Setio et al., 2017) | A dataset containing lung scans for normal and lung nodule classification. |
| | LIDC-IDRI (Armato et al., 2011) | A large dataset containing lung scan for the diagnosis and screening of thoracic cancer. |
| **MRI-scan** | MRNet (Soares & Angelov, 2020) | A dataset collated at the Stanford medical center, containing about 1400 knee MRIs. |

*Corresponding Author*: Tel; +234-811-940-0230
Email Address: emerald.henry@stu.cu.edu.ng

| | | |
|---|---|---|
| **X-ray** | Shenzen dataset (Jaeger et al., 2014) | A dataset collated in Shenzen China on tuberculosis. |
| | COV19 chest x-ray (Maguolo & Nanni, 2019) | A dataset contain chest x-rays for classifying covid-19 from bacterial pneumonia. |
| | Montgomery chest x-ray (Jaeger et al., 2014) | A dataset collated in the Montgomery county. |
| | CXR Images (Kermany, Daniel; Zhang, Kang; Goldbaum, n.d.) | A dataset containing OCT and chest X-rays. |
| | COVIDx (Linda Wang et al., 2020) | A 3-class dataset for normal, viral pneumonia and covid-19. |
| | BIMCV-COV19+ (Vayá et al., 2020) | This is a database of chest x-rays and CT images. |
| | Extensive-XR-CT (X. Yang et al., 2020) | A dataset of CT and X-rays for patients with and without covid. |
| | Posterior-Anterior Chest Radiography COV19 (Haghanifar et al., 2022) | This is a combination dataset of about 15 smaller chest X-ray datasets. |
| **Fundus Images** | Color Fundus (Hajeb Mohammad Alipour et al., 2012) | This Dataset contains fundus images in DR-grading. |

| SEGMENTATION | | |
|---|---|---|
| **Modality** | **Dataset** | **Description** |
| **CT-scans** | IMDTD-18 (Kermany et al., 2018) | A dataset containing about 9000 OCT scans. |
| | Kits19 (Heller et al., 2022) | A dataset annotated for the segmentation of renal tumor. |
| **MRI-scans** | M&MS-21 (Campello et al., 2021) | A dataset containing 375 annotated cardiac magnetic resonance images. |
| | MR-Brain-S (Mendrik et al., 2015) | A dataset containing 20 annotated multi sequence brain MRI. |
| | ERI (Stirrat et al., 2017) | A dataset containing 375 cardiac MRIs. |
| | CHAOS (Kavur et al., 2021) | An MRI annotated dataset for abdominal organ segmentation: kidney and liver. |
| | UKBB (Sudlow et al., 2015) | A dataset for identifying the causes of a wide range of complex diseases. |
| | BrATS-20 (Bakas et al., 2017) | A dataset for brain tissue segmentation, with about 2000 images. |
| | Iseg-17 (Li Wang et al., 2019) | A dataset for brain tissue segmentation consisting of 20 images. |
| **X-ray** | OAC (Peterfy et al., 2008) | A dataset containing annotated knee images. |

*Corresponding Author*: Tel; +234-811-940-0230
Email Address: emerald.henry@stu.cu.edu.ng

| | | |
|---|---|---|
| | DICRLN (Shiraishi et al., 2000) | A database of annotated chest images for detecting lung nodule. |
| | IN-breast (Moreira et al., 2012) | A database containing annotated breast mammograms. |

## 3.2 Classification and Segmentation

Classification of medical images is a vital task in healthcare because of the increasing need for diagnosis, identification and distinction of healthcare images and relevance in intended applications. Varieties of transformer architectures have been employed in modelling with the aim of achieving higher efficiency than previously obtained. Researchers have employed pure vision transformers and hybrid models of disparate architectures for classification of medical images in literature, hence the extensive review of the various transformer-based methods developed for the classification of medical images, with a focus on the image modality, as presented in the writeup.

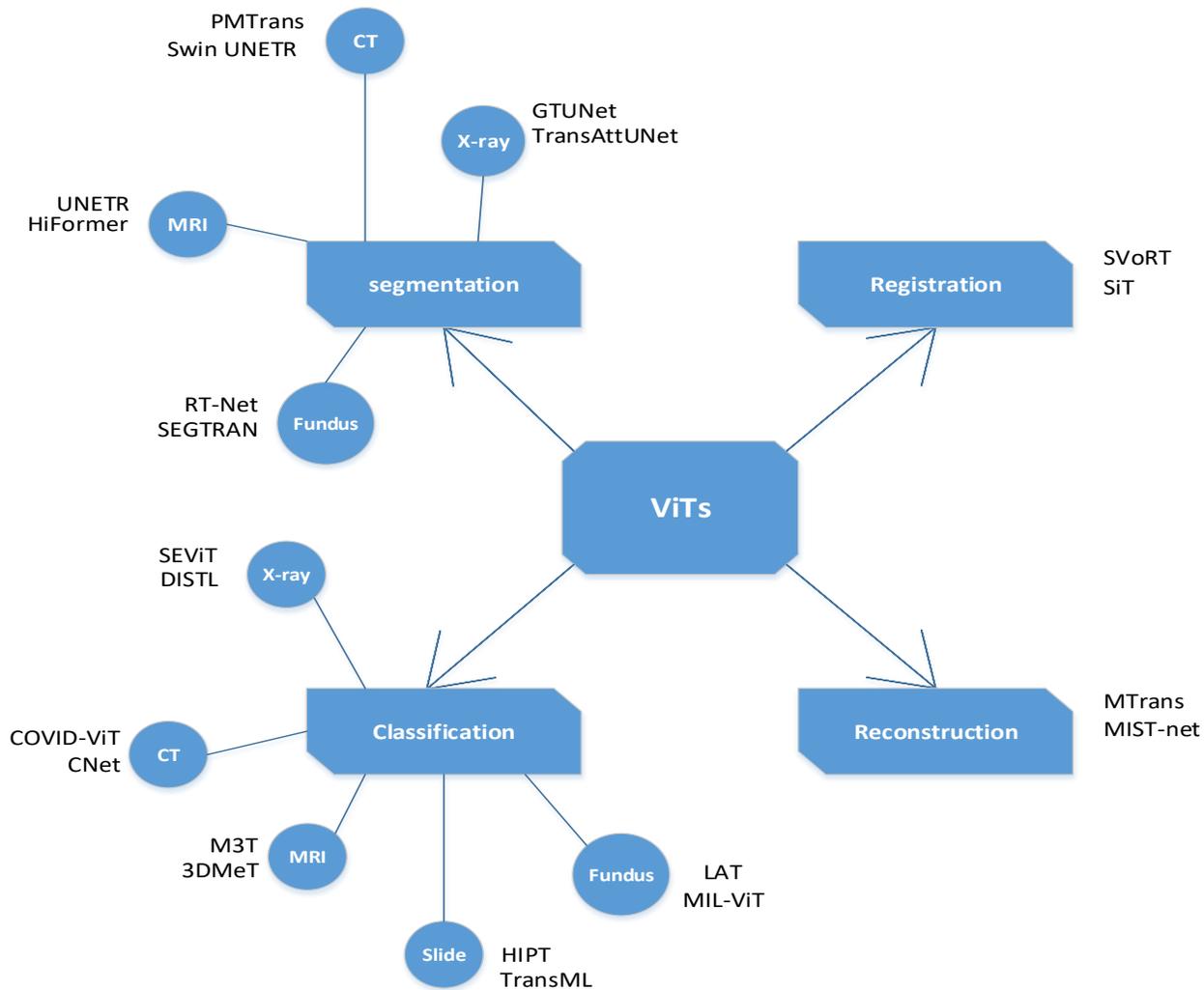

**Figure 3:** A taxonomy on the application of vision transformers in medical imaging detailing prominent models of varying modalities and disparate tasks.

### 3.2.1 Histopathological Images

Various staining substances and methods have aided the diagnosis, detection and classification of tumors and carcinomas in pathology. These slide images, in digital form, serves as a resource bank for the development of computational models to perform automatic diagnosis, to facilitate efficient and faster diagnosis. The earliest known ViT utilized for histopathological image classification, in literature is TransMIL (Shao et al., 2021), which integrate transformers to a multiple instance learning (MIL) framework, in order to introduce correlation between various instances. In this work, MIL is achieved by pooling operations performed on learning instances extracted by a pre-trained CNN (He et al., 2016). These instances, after squaring, are passed into a block containing multi-head self-attention (MSA), positional encoding and an MLP for weakly supervised classification.

*Corresponding Author*: Tel; +234-811-940-0230
Email Address: emerald.henry@stu.cu.edu.ng

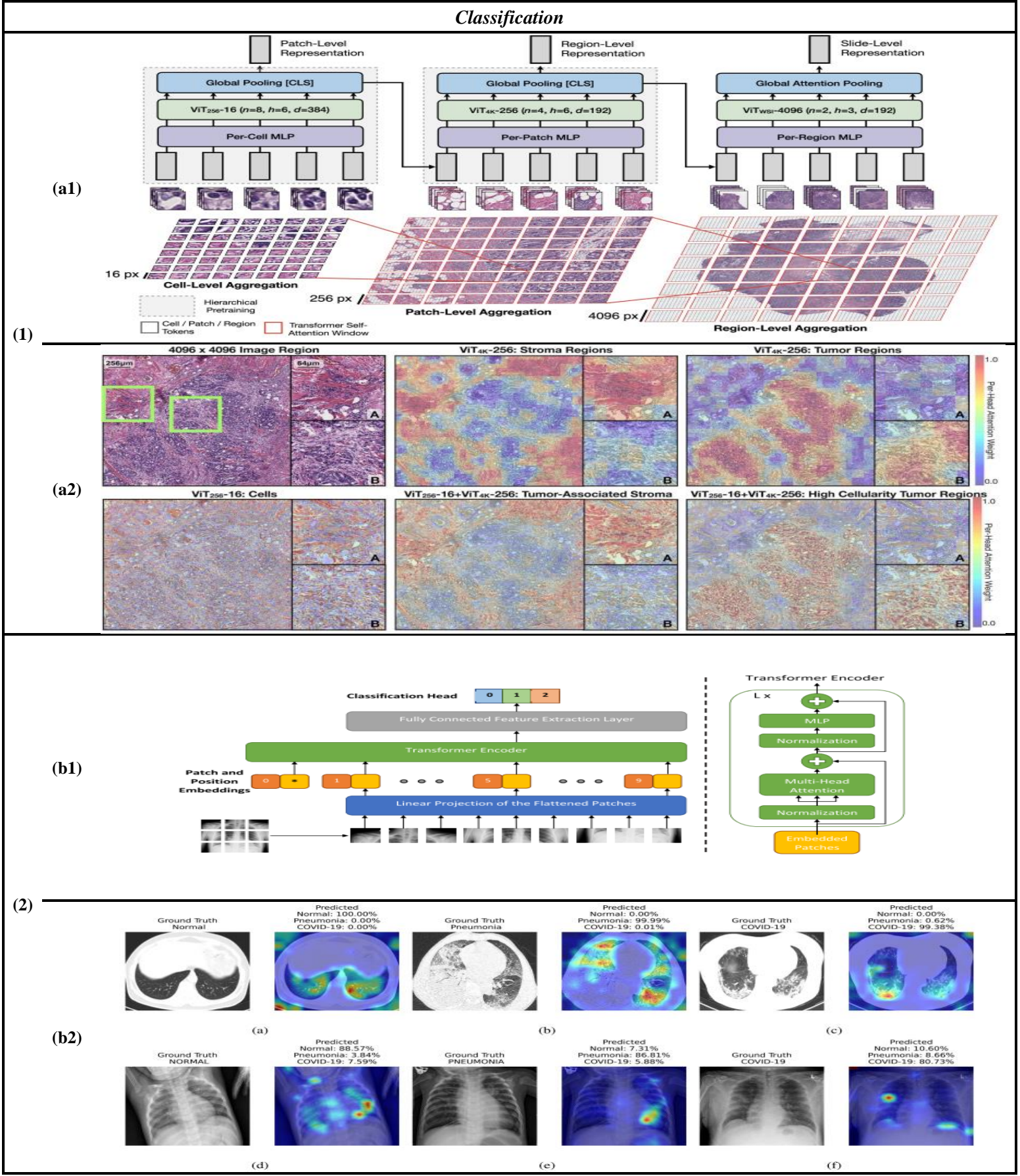

**Figure 4:** (1) Hierarchical Image Pyramid Transformer designed to exploit the hierarchical nature of WSIs while utilizing DINO self-supervised pre-training **(Chen et al., 2022):** HIPT architecture (a1), a colorectal cancer attention map from HIPT (a2). (2) xVITCOS utilizing the pure ViT architecture for chest X-ray classification **(Mondal et al., 2022):** (b1) is the ViT architecture, (b2) is the detection output and score.

*Corresponding Author*: Tel; +234-811-940-0230
Email Address: emerald.henry@stu.cu.edu.ng

The resultant models achieved AUC scores of 93.09% and 96.03% in classifying binary tumors and cancer subtypes respectively. Wang et al. (2022) introduced a contrastive learning strategy based on self-supervised learning (SSL). In general, SSL is known to ameliorate the annotation of large-scale data by utilizing an unlabeled dataset in the extraction of useful representations that generalizes well to multiple downstream tasks. Semantically relevant contrastive learning (SRCL), unlike traditional contrastive learning, seeks to acquire more representations that are informative by aligning closely related positive pairs. This is achieved by pre-training CTransPath (an integration of CNNs and multiple swin transformers) with a plethora of unlabeled data, to serve as a feature extractor for the downstream task performed by SRCL. GasHis-Transformer (Chen et al., 2022) introduces an approach for gastric histopathological image classification based on global and local information capture. Local information capture is realized by a CNN architecture while global information capture is accomplished by attention blocks coupled to CNNs. Advocates of self-supervised ViTs in histopathology (Chen & Krishnan, 2022) proposed a novel vision transformer based on the Hierarchical nature of WSIs. Hierarchical image pyramid transformer (HIPT) (Chen et al., 2022), employs a two-level self-supervised learning framework to exploit both the hierarchical nature of WSIs and the large sequence length of WSI tokens as a result of their pixel nature. This is performed by aggregating visual tokens at three levels (cell, patch and region) in order to form the slide representation and utilizing self-attention as a permutation-equivariant aggregation layer; the HIPT with hierarchical pre-training yielded better performance than SOTA methods for survival prediction and cancer subtyping.

### 3.2.2   Computed Tomography Scans

CT-scans can create detailed cross-sections of bones, soft tissues, organs and blood vessels, these sections can also be formatted to multiple frames that aid in the generation of three-dimensional images. The application of ViTs in computational tomography has largely been focused on thoracic diseases because of the contrast between gas and tissue. These ViTs are designed for two-dimensional, as well as three-dimensional CT images. CTNet as proposed by Liang et al. (Liang et al., 2021) is a hybrid model comprising a CNN feature extractor and ViT for the detection of COVID-19 from three-dimensional chest scans. After training on the COV19-CT dataset, it achieved 88% in F1 evaluation. Barhoumi et al. (2021) proposed scopeformer, a hybrid model for the classification of intracranial hemorrhage from CT images. Their work demonstrates that by stacking several Xception CNN blocks and aggregating their feature maps, a feature-rich map can be developed to serve as the input to a ViT and improve the model performance. Consequently, the model achieved a test accuracy of 98.04% on the corresponding classification task.

The effective diagnosis of pancreatic cancer from two-dimensional CT images was demonstrated by Xia et al (2021). In this work, a region-of-interest (ROI) feature map is developed from annotated training data based on the U-net segmentation algorithm. This pancreas ROI feature map is fed as input to a transformer that is built upon the U-net algorithm for segmentation. The model is trained on a 3-class labeled data for classification, and achieved specificity and sensitivity of 95.2% and 95.8% respectively after training on a dataset of 1321 patents. Swin UNETR (Tang et al., 2021) is a prominent model for the classification of 3 dimensional CT images. It implements a hierarchical encoder for self-supervised pre-training, it was fine-tuned for classification and segmentation tasks at the Beyond the Cranial Vault (BTCV) challenge. Their model outperformed all other models submitted for the challenge.

*Corresponding Author*: Tel; +234-811-940-0230
Email Address: emerald.henry@stu.cu.edu.ng

**Table 2: A summary of the reviewed transformer-based approach for medical image classification.**

| Archi | Modality | Organ | Type | Evaluation | Highlights | Reference |
|---|---|---|---|---|---|---|
| Hybrid | Pathology | Prostate | 2D | Accuracy | The classification task is performed according a grading system regarded as 'Gleason' | Ikromjanov (Al., 2022) |
| Hybrid | Pathology | Stomach | 2D | Precision, F1 | Classification based on the parallel structure of LIM and GIM modules. | Chen et al. (H. Chen et al., 2022) |
| Hybrid | Pathology | Cell | 2D | SP, H-mean, F1 | Utilizes transfer learning and an attention based decoder in classifying cervical cells. | Zhao et al. (C. Zhao et al., 2022) |
| Hybrid | Pathology | Multiple | 2D | Accuracy, AUC, F1 | Introduces a TAE module that aid the ViT in aggregating tokens, which are passed into an FFN. | TransPath (Xiyue Wang et al., 2022) |
| Hybrid | Pathology | Multiple | 2D | Accuracy AUC | A CNN encoder attached to a ViT decoder for the extraction of spatial information from WSIs | TransML (Shao et al., 2021) |
| Hybrid | Pathology | Multiple | 2D | Accuracy, Precision, recall | A CNN, transformer encoder to capture high level features for classification. | i-ViT (Z. Gao et al., 2021) |
| Hybrid | Pathology | Multiple | 2D | MSE | Combining self-supervised learning and transfer learning in extracting morphological features from WSIs. | R. J. Chen et al. (R. J. Chen & Krishnan, 2022) |
| Pure ViT | Pathology | Multiple | 2D | AUC | Developed a novel Hierarchical transformer that leverages self-supervised learning for classifying cancer and survival prediction. | R. J. Chen et al. (R. J. Chen et al., 2022) |
| Hybrid | Pathology | Lung | 2D | ACC, PRE, SE, SP, RE. | Consisting of a CNN feature extractor and a feed forward network for classifying WSIs. | GTN (Y. Zheng et al., 2022) |
| Pure ViT | CT | Chest | 3D | Accuracy, F1 | Defined a new ViT architecture that was implemented for covid-19 classification. | COVID-ViT (X. Gao., 2021) |
| Hybrid | CT | Chest | 3D | F1 | Proposed a two-stage process of segmentation using U-net then classification using swin-transformer. | Zhang et al. (L. Zhang & Wen, 2021) |
| Hybrid | CT | Chest | 3D | Accuracy, Precision, Recall, F1 | The implementation of Wilcoxon signed-rank test for preserving CT slices after which spatial features are extracted by a transformer that contains convolutions. | Hsu et al (Hsu et al., 2021) |
| Hybrid | CT | Pancreas | 3D | Sensitivity, Specificity, AUC | Localization is achieved by U-net and then passed to a transformer. | Xia et al. (Y. Xia et al., 2021) |
| Pure ViT | CT | Lung | 2D | F1 | Utilized a teacher student framework to aid knowledge distillation. | Li et al. (Jingxing Li et al., 2021) |
| Hybrid | CT | Brain | 2D | - | The stacked several feature maps produced by a CNN into a ViT to improve classification accuracy. | Scopeformer (Barhoumi & Ghulam, 2021) |
| Hybrid | CT | Lung | 3D | F1 | Combines convolution and attention in the classification of 3D chest images. | CNet (Liang et al., 2021) |
| Pure ViT | CT | Chest | 2D | Precision, Recall, F1, Specificity | Introduces a transfer learning approach of multiple stages where ViT performs the upstream task. | xViTCOS (Mondal et al., 2022) |
| Pure ViT | CT | Multiple | 3D | specificity | Introduced a new 3D transformer that has a hierarchical encoder for SSL. | Tang et al. (Tang et al., 2021) |
| Pure ViT | MRI | Ear | 3D | Accuracy, Precision | This work contains the first transformer model for multi-modal image classification performed on ear MRI. | Matsokus et al. (Y. Dai et al., 2021) |

*Corresponding Author*: Tel; +234-811-940-0230
Email Address: emerald.henry@stu.cu.edu.ng

| | | | | | | |
|---|---|---|---|---|---|---|
| Hybrid | MRI | Hepatic | 2D | MAE | This work utilizes 3 parallel CNN encoders for feature extraction and a transformer decoder for classification of hepatocellular carcinoma. | mfTrans-Net (J. Zhao et al., 2021) |
| Hybrid | MRI | knee | 3D | - | They introduced 3D convolutional block encoding to reduce cost of computation and utilized a teacher student approach to train ViT. | 3DMeT (Sheng Wang, 2021) |
| Hybrid | MRI | Brain | 2D | MAE, PC | Utilizes two parallel CNNs; one extracts features from the whole image, the other from patches. A transformer acts as the decoder. | He *et al.* (S. He et al., 2022) |
| Hybrid | MRI | Alzheimer | 3D | Accuracy, AUC | This work proposes an architecture that combines a 2D and 3D CNN for classification, with a transformer encoder. | M3T (Jang & Hwang, n.d.) |
| Pure ViT | MRI | Brain | 2D | Accuracy | The work pre-trains a ViT on ImageNet and fine-tunes it for brain tumor classification. | Tummala *et al.* (Tummala, 2022) |
| Pure ViT | MRI | Brain | 3D | AUC, PC | Proposes an attention based encoder requiring patch meshes, for classification tasks. | Dahan *et al.* (Dahan et al., 2022) |
| Hybrid | MRI | Brain | 3D | AUC, MAE | Pre-trains a hybrid model in a self-supervised manner for 3D brain disease diagnostic etc. | Jun *et al.* (Jun et al., 2021) |
| Hybrid | MRI | Brain | 2D | R2, MAE, CC | This work presents an end-to-end attention guided deep learning approach for gestational age predication utilizing an attention based intermission. | Shen *et al.* (Shen et al., 2022) |
| Hybrid | X-ray | Lung | 2D | Accuracy | Develops a hybrid model consisting of CNNs for image processing and a multiple swin-transformers in sequence for classification. | Sun *et al.* (Sun & Pang, n.d.) |
| Hybrid | X-ray | Lung | 2D | - | Utilizes a CNN encoder and incorporates average pooling to the attention block MLP. | H. Xu *et al.* (H. Xu et al., 2022) |
| Pure ViT | X-ray | Lung | 2D | AUC | Introduces a ViT with random patch distribution for multi-task learning. | p-FESTA (S. Park & Ye, 2022) |
| Pure ViT | X-ray | Lung | 2D | AUC | This work introduces a self-ensemble ViT architecture to improve ViT robustness. | SEViT (Almalik et al., 2022) |
| Pure ViT | X-ray | Lung | 2D | AUC | Demonstrated split performance without adulterations to performance. | Park *et al.* (S. Park et al., 2021) |
| Hybrid | X-ray | Lung | 2D | AUC | Proposes an approach of diagnosis through knowledge distillation. | DISTL (S. Park et al., 2022) |
| Pure ViT | Fundus | Eye | 2D | Accuracy, F1, Recall, Precision | Involves a transformer pre-training on a large amount of fundus dataset then fine-tune to classification task. | MIL-ViT (Shuang Yu, 2021) |
| Hybrid | Fundus | Eye | 2D | AUC, Kappa | Investigates lesions as a localization problem of a weakly supervised nature, for classifying diabetic retinopathy grades and diagnosing lesions. | LAT (Rui Sun, 2021) |
| Pure ViT | Fundus | Eye | 2D | AUC | Investigates the performance of DEiT compared to SOTA CNN for classification tasks. | Matsoukas *et al.* (Matsoukas et al., 2021) |

*Corresponding Author*: Tel; +234-811-940-0230
Email Address: emerald.henry@stu.cu.edu.ng

The segmentation of CT-scans is important in the diagnosis, evaluation and monitoring of various phenomena in healthcare. Swin UNETR (Tang et al., 2021) also performs 3D segmentation primarily for computed tomography images. It attaches self-supervised heads, aggregated by contrastive learning strategy to a swin transformer encoder. It also achieved state-of-the-art performance in segmentation. Similarly Xia *et al* (2021) hybrid model train by classification supervision can be utilized for CT segmentation tasks as well, and for CT classification task as discussed above. Because of the computational cost involved in modelling global representation on full resolution images Zhang (2021) proposed PMTrans, a hybrid model that incorporates mulit-scale attention to a CNN feature extractor with the aim of capturing diverse range relationships from multi-resolution images. This method is recorded to have outperformed other CNN and Transformer based models at the time of publication. UNETR (Tang, et al., 2022) was designed for semantic segmentation of 3D brain MRI and spleen CT scans. It attaches a transformer encoder to a Unet-like CNN decoder for semantic segmentation.

### 3.2.3 Magnetic Resonance Imaging

MRI provide very detailed anatomical images due to its powerful and effective non-invasive imaging technology: it also produces two-dimensional as well as three-dimensional images. It is therefore of importance to consider the dimensionality of the image instance that the modelling method is suited. M3T (Jang & Hwang, 2022) is a three-dimensional image processing technique for the classification of Alzheimer's disease. It proposes two- and three-dimensional representation with pre-trained 2D CNN and 3D CNN respectively. These CNNs are designed to append inductive bias to the modelling technique while global sequential information from multiple planes are captured by the ViT attached sequentially to the 2D-3D CNNs structure. Their model was trained on the Alzheimer's Disease Neuroimaging Initiative (ADNI) dataset and was validated on the Australian Imaging, Biomarker and Lifestyle Flagship Study of Ageing (AIBL) dataset. The model outperforms all other models in Area under Curve (AUC) evaluation. Tummala (2022) employs an ensemble of ViT models pre-trained and fine-tuned on ImageNet for brain tumor classification, demonstrating the efficiency of disparate domain information transfer. They achieved a test accuracy of 98%. Works on age prediction based on MRI were performed by He *et al.* (2022) and Shen *et al.* (2022). On the one hand, the former attempts at estimating brain age by a global-pathway-local-pathway transformer network that aids the simultaneous capture of global and local representations from images. The particular model was trained on a collective of six publicly available dataset. After evaluation on two additional datasets, it exhibited state-of-the-art performance in age prediction. On the other hand, the latter attempted at extracting age-specific morphological information from Fetal MRI to promote age-based classification. They also demonstrate the inferiority of traditional CNN approaches to that of attention-guided CNN methods: achieved by incorporating attention-guided mask inference to a traditional ResNet-50 architecture. Their method resulted in $R^2$ score of 0.945 and MAE of 6.7 days.

The fully convolutional transformer (Tragakis et al., 2022) leverages the success of U-net algorithm in development of a fully-convolutional, depth-wise transformer that replaces the convolution blocks in U-net architecture with fully convolutional transformers. This method aids efficient segmentation that considers the fine-grained nature of medical images. Their model extracts global information as well as capturing hierarchical attributes from features. UNetFormer (Xu, et al., 2022) attempts to adapt the dubbed UNetFormer architecture by redefining each encoder block with swin transformers and each decoder block with CNNs; they also included skip connections at different data resolutions. For encoder pre-training, they incorporated self-supervised learning with the aim of letting the model randomly predict masked volumetric tokens.

*Corresponding Author*: Tel; +234-811-940-0230
Email Address: emerald.henry@stu.cu.edu.ng

*Segmentation*

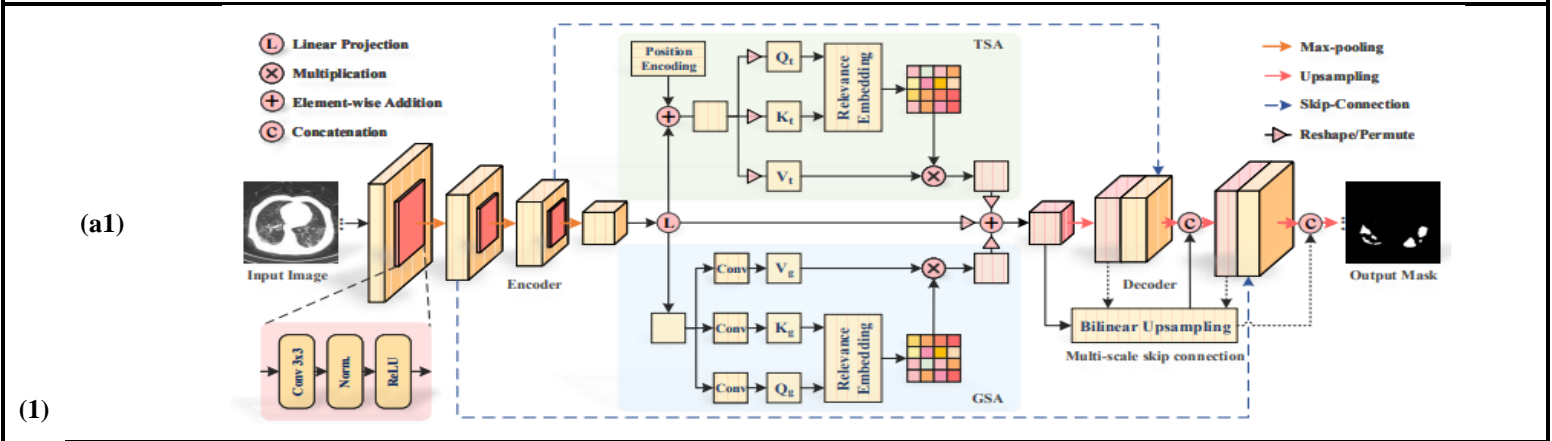

(1) (a1)

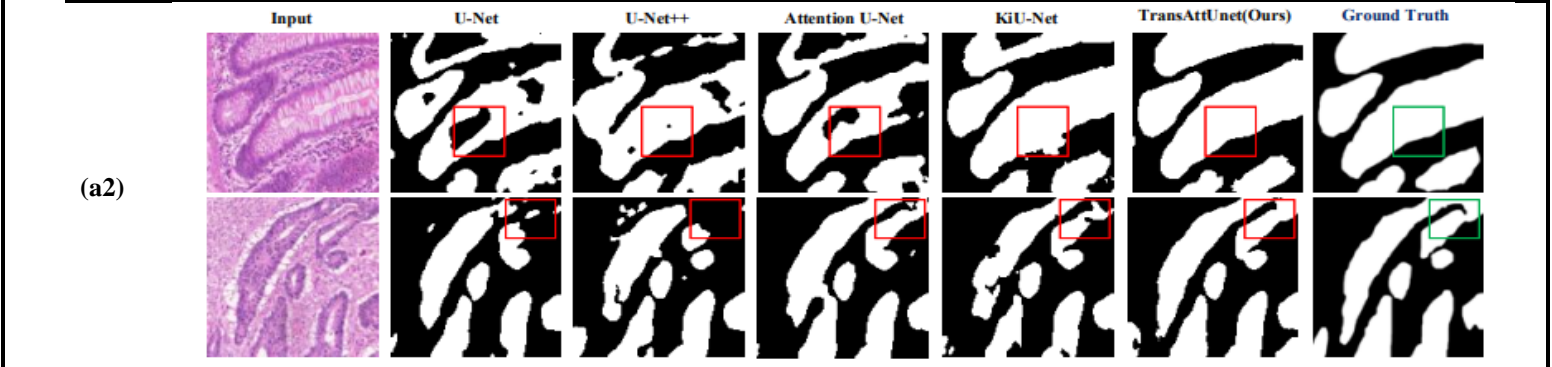

(a2)

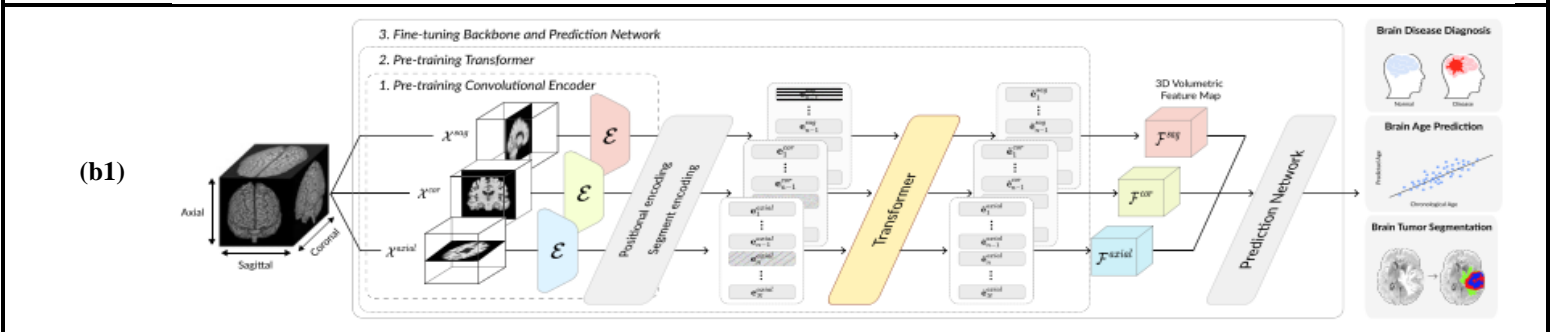

(b1)

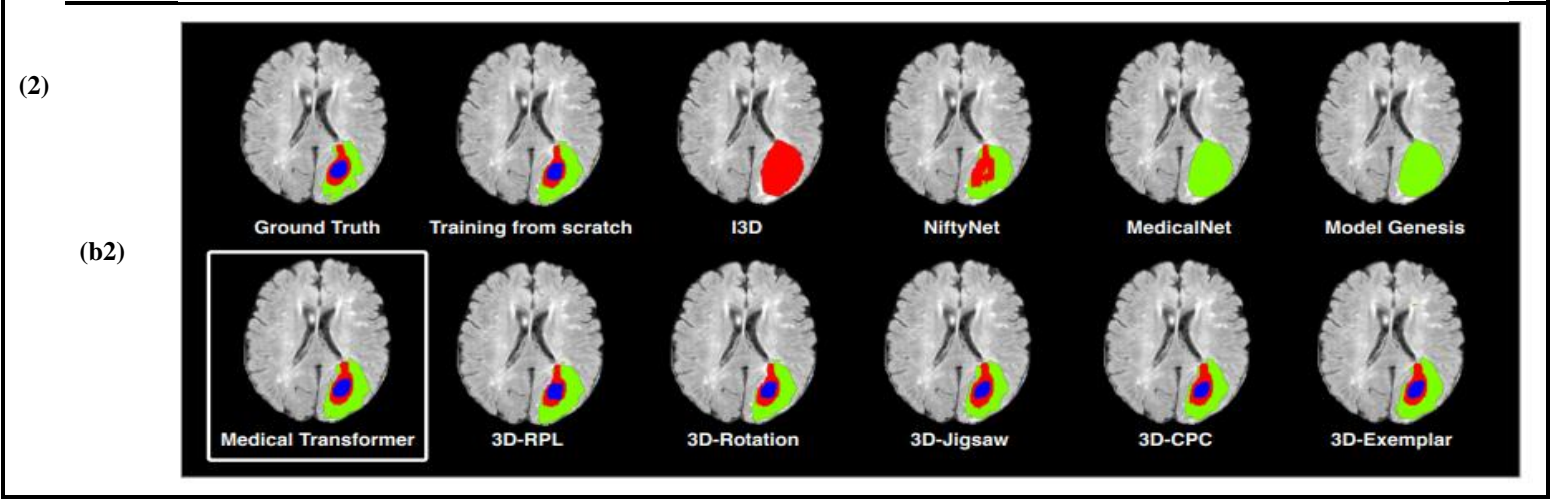

(2) (b2)

**Figure 5:** (1) TransAttUnet (Chen et al., 2021) is an attention guided u-net with transformers for segmentation: its architecture (a1) utilizes TSA and GSA mechanism, (a2) is a comparative gland segmentation result. (2) medical transformer (Jun et al., 2021) is utilized for 3D MRI analysis: (b1) is its architecture that shows that slices are processed in 2D before being combined to form the output, (b2) is a comparative between medical transformer and other approaches.

*Corresponding Author*: Tel; +234-811-940-0230
Email Address: emerald.henry@stu.cu.edu.ng

An alternative modelling method that proposes a deformable bottleneck transformer module to aid the capture of shape; TransBTSV2 (Jiangyun Li et al., 2022) retains a CNN encoder decoder structure with only major changes to the bottleneck region, it requires no pre-training and stores local information through its CNN encoder while also capturing global sequential representations. This architecture was tested on four publicly available datasets and generated results that are comparable with state-of-the-art classification techniques. Another variation of this approach is the nnFormer Zhou (2021), attempts to aid the capture of volumetric information by defining a local and volume based bottleneck on a conventional CNN encoder-decoder structure. It achieved significant reduction of both the dice score and Haudroff distance metric.

Not a lot of researchers apply major changes to their transformer structure. However, in an attempt to reduce the dimensional and computational complexity of the vanilla transformer Xie *et al.* (2021) developed CoTr. This deformable transformer unlike the vanilla transformer that pays unequal attention to all image positions. Utilizing the deformable transformer coupled to a CNN encoder-decoder, they effectively processed multi-scale and high-resolution feature maps while capturing local image representations.

Another modelling approach is to convert three-dimensional images to two-dimensional slices (Jun et al., 2021), utilizing only two-dimensional convolutions. This network is pre-trained using DINO self-supervision, after which the two-dimensional slices are re-combined for the purpose of prediction. This method has been successfully applied for regression and classification, as well as segmentation tasks.

### 3.2.4    X-ray

The passage of X-rays through a body can aid the generation of informative images of tissues and the internal structure of the body. X-rays are inexpensive and convenient and have a wide range of application in medical imaging such as cancers, fractures and various pneumonia (including COVID-19) etc. Thus, they are an important source of medical data capture to clinicians and researchers alike. SEViT (Almalik et al., 2022) attempts to model self-ensemble transformers based on experiment that proved that the feature representations learned by the initial ViT blocks are generally unaffected by adversarial perturbations. In order to model resilience to adversarial attacks, they proposed learning multiple classifiers that aggregates feature representations learned from initial ViT blocks to those learned from final ViT blocks. This method leverage the details presented from the final ViT representation as well as the robustness of intermediate ViTs. Mondal *et al* (2022) employed a pure transformer for the detection and classification of COVID-19 from X-ray images. In their work, a multi-stage transfer learning strategy based on ViT is adopted where a ViT is pre-trained on domain specific data for the extraction of domain relevant information. This feature extractor is attached to a conventional transformer that performs classifications based on its fully connected multi-layer perceptron (MLP). Chest X-ray data and other X-ray forms required for adequate modelling are scarce. However, availability of such data is on the rise. With an increase in the available data comes increasing demand for annotation. Park (2022) tried to provide solutions to these issues in his works p-FESTA (Park & Ye, 2022) and DISTL (Park et al., 2022). For scarcity of training samples, they proposed p-FESTA that employs multi-task distributed learning that is based on federation and shared learning. They achieved this by exchanging the CNN in the original Federated Split Task Agnostic (FESTA) with random patch permutation with an aim of improving performance of the multi-task learning while maintaining privacy. For the problem of annotating newer datasets, they proposed DISTL that incorporates self-supervision and self-training to ViTs with the aim of developing a model that can learn useful representations from un-annotated data. DISTL was evaluated on data from three hospitals and improved in performance with increase in the size of unlabeled training data.

*Corresponding Author*: Tel; +234-811-940-0230
Email Address: emerald.henry@stu.cu.edu.ng

Table 3: A summary of the reviewed transformer-based approach for medical image segmentation.

| Archi | Modality | Organ | ViT Enco\|inter\|deco | Type | Evaluation | Highlights | Reference |
|---|---|---|---|---|---|---|---|
| Hybrid | CT | Multiple | 1\|0\|0 | 3D | Dice, HD | Introduced a new 3D transformer that has a hierarchical encoder for SSL. | Swin UNETR (Tang et al., 2021) |
| Hybrid | CT | Pancreas | 0\|0\|1 | 3D | Sensitivity, Specificity, AUC | A model comprising u-net and transformer is trained based on classification and segmentation supervision. | Y. Xia *et al.* (Y. Xia et al., 2021) |
| Hybrid | CT | Multiple | 0\|0\|1 | 2D | Dice | Aimed at reducing the computational cost of current ViTs by limiting learning to the capture of multi-range relationship between varying resolutions. | PMTrans (Zhuangzhuang Zhang, 2021) |
| Hybrid | CT, MRI | Brain, Spleen | 1\|0\|0 | 3D | Dice, HD | A hybrid transformer comprising a transformer encoder and a CNN decoder for brain and spleen classification. | UNETR (Hatamizadeh, Tang, et al., 2022) |
| Hybrid | MRI, CT, | Multiple | 1\|1\|1 | 3D | sensitivity | A U-net shaped, depth wise, fully convolutional transformer designed for medical image segmentation. | HiFormer (Heidari et al., 2022) |
| Hybrid | MRI, CT | Liver | 1\|1\|1 | 3D | Dice, HD | Comprising convolution and transformer in each block of the depth-wise architecture. | UNetFormer (Hatamizadeh, Xu, et al., 2022) |
| Hybrid | MRI, CT | Multiple | 1\|0\|1 | 3D | Dice | Incorporates bidirectional multi-head attention to a depth-wise U-net structure. | MedFormer (Y. Gao et al., 2022) |
| Hybrid | MRI, CT | Multiple | 0\|1\|0 | 3D | Dice, HD | Utilizes a CNN encoder-decoder structure with a transformer bottleneck. | TransBTSV2 (Jiangyun Li et al., 2022) |
| Hybrid | MRI, CT | Multiple | 0\|1\|0 | 3D | Dice, HD | Introduces a local and volume type of attention mechanism that aids in learning global representations. | nnFormer (H.-Y. Zhou et al., 2021) |
| Hybrid | CT, MRI | Multiple | 0\|1\|0 | 3D | Dice | Introduces deformable self-attention to aid in processing multi-scale, high-resolution images. | CoTr (Xie, Zhang, Shen, et al., 2021) |
| Hybrid | MRI | Brain | 0\|0\|1 | 3D | Dice | Converts a 3D image into two-dimensional slices, processes image in 2D then recombines as output. | Medical Transformer (Jun et al., 2021) |
| Pure ViT | X-ray, CT | Lung | 1\|1\|1 | 2D | IoU, Accuracy | Utilizes several swin-transformers with patch and positional encoding to achieve classification and segmentation training. | Sun *et al.* (Sun & Pang, n.d.) |
| Hybrid | X-ray | Lung | 0\|0\|1 | 2D | Dice | Introduces a ViT with random patch distribution for multi-task learning. | Park *et al.* (S. Park & Ye, 2022) |
| Hybrid | X-ray | teeth | 1\|0\|0 | 2D | Dice | Utilizes a Fourier descriptor based loss function to aid in integrating the shape after which it is passed to grouped transformer blocks. | GTUNet (Yunxiang Li et al., 2021) |
| Hybrid | CT, X-ray | Multiple | - | 2D | Dice | Aims at solving the information recession issue by attempting multi-level attention. | TransAttUNet (B. Chen et al., 2021) |
| Hybrid | Fundus | Eye | 0\|1\|0 | 2D | Precision, Recall, AUC | Defined two kinds of transformer blocks: global and relation transformers in order to aid the detection of minute sizes and blurred borders. | RT-Net (S. Huang et al., 2022) |
| Hybrid | Fundus | Multiple | - | 2D, 3D | Dice | Utilizes squeeze and expansion blocks that serves to regularize the self-attention module as well as learn diversified representations. | Segtran (S. Li et al., 2021) |

*Corresponding Author*: Tel; +234-811-940-0230
Email Address: emerald.henry@stu.cu.edu.ng

A lot of the research on the X-ray modality is performed for classification however, there are a few works that attempts the segmentation of chest x-rays. Sun *et al.* (Sun & Pang, n.d.) Developed a model for the classification and segmentation of chest X-rays based solely on the swin transformer. So far, very few models employ just the transformer architecture in performing segmentation tasks. Yet they achieved a segmentation accuracy of 95% from three variations of their swin-transformer model. The p-FESTA method discussed earlier also tested for segmentation of X-rays; inducing federation and shared learning towards the performance of segmentation tasks after self-supervised pre-training. TransAttUnet (Chen et al., 2021) was developed for the accurate segmentation of organs and lesions from X-ray and CT imaging by defining an encoder-decoder network with multi-scale skip connection for processing higher resolution images and multi-level guided attention for mapping global relationships between multiple resolutions. These skip connections were also applied to the decoder part of the architecture to aggregate semantic-scale up-sampling features, alleviating information recession and developing a more detailed pixel map. To aid in root canal therapy assessment Li *et al.* (2021) developed GT U-Net that retains the depth-wise nature of U-net but replaces the convolutions with group transformer-hybrids. The idea behind implementing group transformers rather than individual vanilla-type transformer is to reduce computational cost. They also defined a shape-sensitive Fourier Descriptor loss function in order to aid model optimization.

### 3.2.5 Fundus

Fundus photography aids in the diagnosis of a variety of medical conditions. Due to its complex nature; comprising retina, macula, optic disc, fovea and blood vessels, it is expedient that computation be employed in more efficient and reproducible diagnostics. Rui *et al.* (2021) developed an encoder-decoder decoder structure with a pixel relation based encoder and a filter based decoder. This model employs weakly supervised training via filter based transformer decoder, as well as lesion region importance and lesion region diversity to enable the model learn filters well. The model was tested for DR grading and lesion discovery, and recorded state-of-the-art performance.

With the aim of aiding ophthalmologist in the automatic segmentation of diabetic retinopathy lesions Huang *et al* (2022) developed RT-Net following a clinical approach: They investigated the pathogenic causes of diabetic retinopathy lesions and found that certain lesions present relative patterns with each other and appear close to specific vessels. This finding aided the proposition of a relation transformer block composed of self-attention for global relationship and cross-attention that enables interactions between lesion and vessel features. They also proposed a global transformer block to aid the capture of finer details of small lesion patterns. Their dual-transformer approach was capable of segmenting four kinds of DR lesions and attains state-of-the-art performance. Segtran (Li et al., 2021) was developed with the aim of capturing fine details as well as global features simultaneously. They successfully developed a transformer-based approach that has unlimited effective receptive field at high and low resolutions by utilizing squeeze and expansion transformer networks, with each transformer block performing a unique function: the squeeze block attempts regularization of the attention block to aid global information capture, the expansion block learns a diverse array of representations. Tested on the BraTS dataset (Bakas et al., 2017) the model achieved the most segmentation accuracy.

*Corresponding Author*: Tel; +234-811-940-0230
Email Address: emerald.henry@stu.cu.edu.ng

**Table 1: A summary of the reviewed transformer-based approaches for medical image registration and reconstruction.**

| Archi | Modality | Organ | Type | Evaluation | Reference |
|---|---|---|---|---|---|
| Hybrid | MRI | Brain | 3D | Proposed slice-to-volume registration that predicts the transformation of a slide based on information from other slides. | SVoRT (Junshen Xu et al., 2022) |
| ViT | MRI | Brain | 3D | Utilizes a pure transformer for the projection of surface data on a curved manifold | SiT (Dahan et al., 2022) |
| Hybrid | MRI, CT | Multiple | 3D | Presents a hybrid transformer for image registration along with 2 model variants. | TransMorph (J. Chen et al., 2021) |
| Hybrid | CT | Multiple | 3D | An encoder-decoder architecture composed of transformer blocks for denoising medical images. | Eformer (Luthra et al., 2021) |
| ViT | CT | Multiple | 3D | Comprises a symmetrical encoder-decoder architecture solely based on transformers for denoising. | TED-net (D. Wang et al., 2021) |
| ViT | MRI | Multiple | 2D | Cascading swin transformers forming a reconstruction network with a self-supervised learning strategy | DSFormer (B. Zhou et al., 2022) |
| Hybrid | MRI | Multiple | 2D | Proposed a CNN-Transformer for both super-resolution and MRI reconstruction | $T^2$ Net (C. M. Feng et al., 2021) |
| ViT | MRI | Multiple | 2D | A multi-modal transformer capable of transmitting features from the target modality to the auxiliary modality. | MTrans (C.-M. Feng et al., 2022) |
| Hybrid | CT | Multiple | 2D | Introduces a flexible architecture for residual data and image capture, introduces enhancement filters for preserving edges, then a swin transformer to aid reconstruction. | MIST-net (Pan et al., 2022) |

## 3.3 *Registration and Reconstruction*

Medical image registration becomes relevant when we intend to analyze the same image captured with different modalities or at different times. Registration aims at establishing relationships between static and moving images by finding dense per-voxel displacement. In recent times, transformers are seen as the architecture of choice for extracting relating features from multimodal images required in registration tasks because of they offer better understanding of spatial representations. SVoRT (Junshen Xu et al., 2022) attempts volume registration from slides based on an attention mechanism. Features extracted by a ResNet architecture, as well as the position of slices in 2D and 3D are fed into the transformer to encode spatial and global representations. SVoRT was tested on real world data and achieved state-of-the-art performance. Dahan *et al* (2022) developed a transformer-based technique for the projection of surface features to a curved surface by leveraging multi-head self-attention and spatial encodings.

The transformation of signals into an interpretable image that can serve as data for diagnosis is a task that is performed efficiently by transformers.

*Corresponding Author*: Tel; +234-811-940-0230
Email Address: emerald.henry@stu.cu.edu.ng

*Registration and Reconstruction*

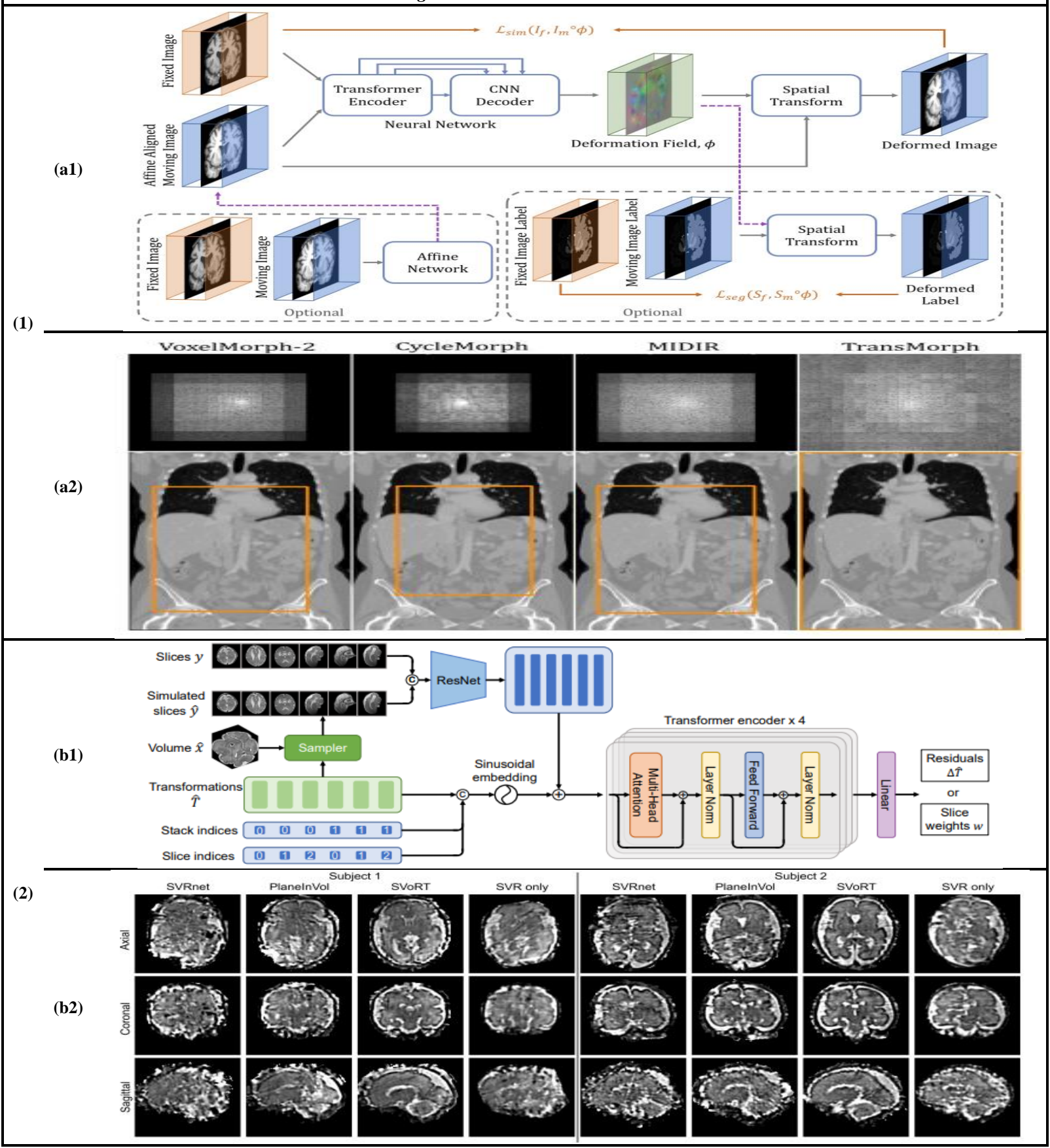

**Figure 6:** (1) TransMorph (Chen et al., 2021) is a model developed for unsupervised medical image registration, its architecture (a1) takes in two inputs and generates a nonlinear warping function. (a2) depicts the ERF of other methods compared to it. (2) SVoRT (Junshen Xu et al., 2022) is an iterative transformer for fetal MRI reconstruction: (b1) depicts its architecture, (b2) shows reconstructed volumes by the model.

*Corresponding Author*: Tel; +234-811-940-0230
Email Address: emerald.henry@stu.cu.edu.ng

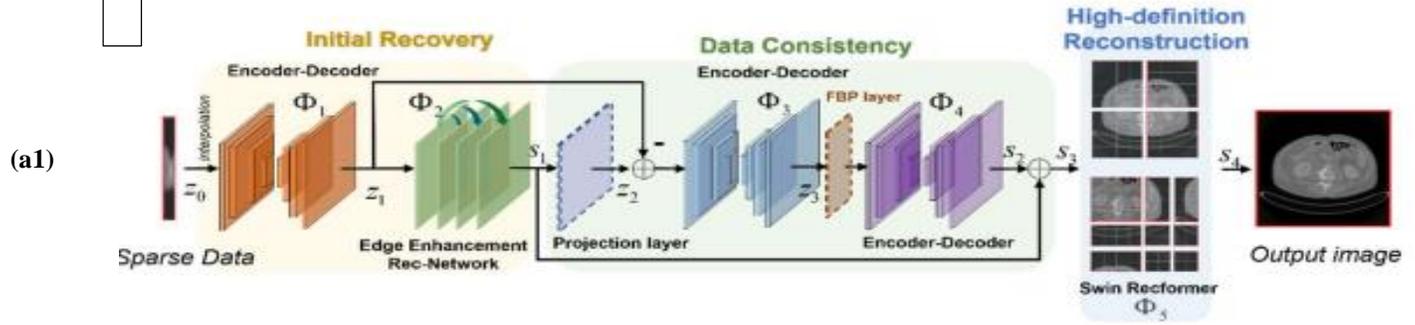

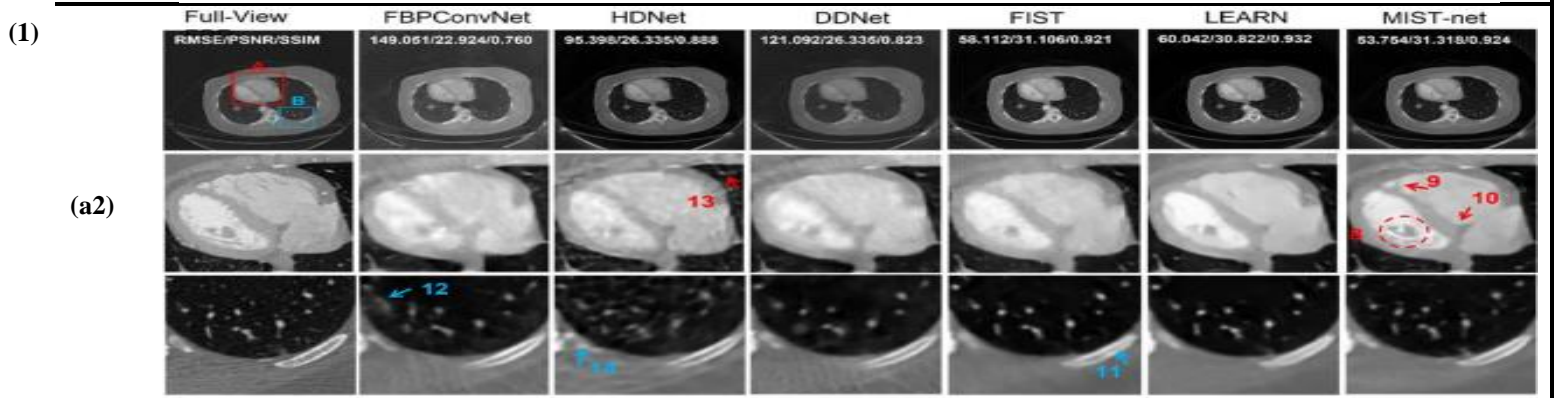

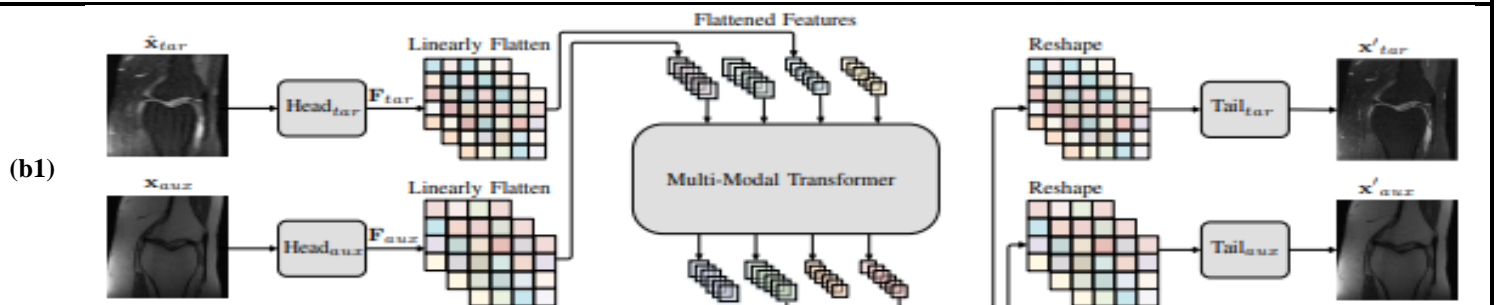

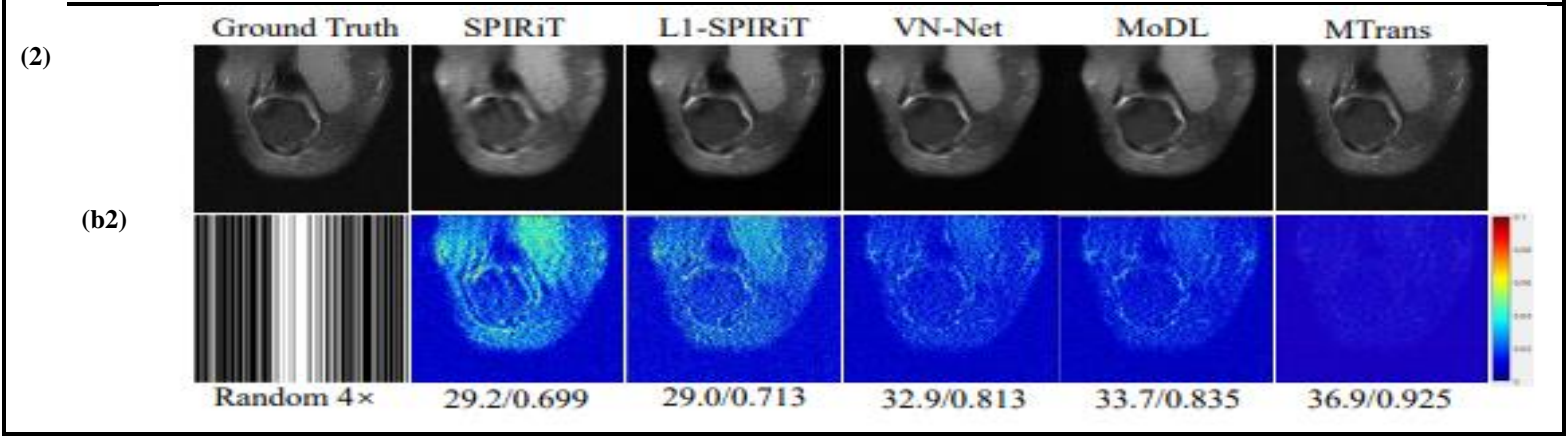

**Figure 7: (1) MINST-net (Pan et al., 2022) is a model based on the swin transformer for sparse-view CT reconstruction; consists of an architecture with 3 stages (a1), (a2) is a comparative of the result of other methods alongside MINST-net. (2) MTrans (Feng et al., 2022) is a multimodal transformer for MRI reconstruction. (b1) depicts its architecture while (b2) shows comparison between different multi-coil reconstruction methods.**

*Corresponding Author*: Tel; +234-811-940-0230
Email Address: emerald.henry@stu.cu.edu.ng

This progress has led to; the reduction of the number of MRI measurements required to establish a scan, aid in reducing the radiation dose required in CT scans, ease of rebuilding surgical scenes. DSFormer (Zhou et al., 2022) presents a self-supervised learning approach based on the transformer and aids the acceleration of multi-contrast MRI reconstruction. It achieves this by developing deep conditional cascade transformer from swin transformers with two unique strategies to encourage information sharing. During evaluation, DSFormer achieved almost equivalence performance when train on full supervision and self-supervision. T$^2$ Net (Feng et al., 2021) developed a hybrid model of convolutions and transformer architecture for MRI reconstruction and super-resolution by situating task transformer networks after two parallel CNNs. Feng *et al.* (2022) presents (MTrans) that utilizes a transformer architecture for multi-modal MR imaging for the purpose of global information capture. They additionally defined a cross-attention module to extract multi-scale information from the different modalities and recorded state-of-the-art performance upon evaluation. MIST-net (Pan et al., 2022) was developed for sparse-view tomographic reconstruction. It entails a robust architecture of three modules; initial recovery based on a flexible network architecture comprising convolution and pooling layers arranged in an encoder-decoder structure; data consistency aimed at preserving edge information by utilizing an enhancement filter defined by convolution and concatenation operations, high-definition reconstruction via swin transformers. This model performed efficiently on evaluation.

# 4 DISCUSSION

Transformer architecture and its application cut across various imaging modalities in medical imaging. However, there hasn't been a clear and detailed comparison between transformers and their CNN counterpart. Comparatively, we consider the performance of both architectures; with the exploration of the strengths and weaknesses as well as on their performance under various scenarios.

## 4.1 Key Properties of Transformers

A few distinctive properties govern the behaviors of transformers. A good understanding of its strengths and drawbacks have informed researchers on ways to deploy transformers for the creation of problem-specific and more efficient models that are relevant in the real world. These properties are discussed below.

### 4.1.1 Long-Range Dependency

In natural language processing (NLP) long-range dependency can be viewed as the preservation of contextual information in a sequence of word tokens, this can be transmitted to computer vision as the relationship between patches of an image (Devlin et al., 2019). This feature can be attributed to the multi-head, self-attention module that maps all token together with a constant distance thereby capturing token-to-token relationships. CNNs lacks such feature hence have limited receptive fields (Kim et al., 2021).

### 4.1.2 Model Capacity

Transformers aggregate projections progressively at a constant scale whereas CNNs aggregate projections through a series of convolution and pooling operations that continuously rescales the image. Constant scale multi-processing aids in better preservation of global information than rescaling operations (Yanghao Li et al., 2022). Similarly, transformers have a better loss landscape than CNNs due to self-attention, and it leads to better generalizability (Park & Kim, 2022).

### 4.1.3 Integration and Manipulation

Due to the dynamic nature of transformers, a variety of architectures, hybrid or pure, can be created. This property is necessary for reducing the limitations of the method. Additionally, their performance improves with training size and model capacity, a property that also induces increase computation cost and time.

*Corresponding Author*: Tel; +234-811-940-0230
Email Address: emerald.henry@stu.cu.edu.ng

### 4.1.4 Adversarial Noises

One disadvantage of CNNs is their vulnerability to adversarial noises, these limit the model's ability to output an ideal representation of the input data (Choi et al., 2022). Transformers are generally more robust to perturbations and corruptions (Bhojanapalli et al., 2021).

### 4.1.5 Inductive Bias

The scale-adjustment processing employed in convolutional neural networks accords them the ability to extract more local information from individual pixels, leading to faster convergence and better performance when trained on smaller datasets (Cordonnier et al., 2019). This is not the case with transformers because same-scale processing in transformers capture more global, than local, information (Ramachandran et al., 2019).

## 4.2 Transformer and CNNs

Considering the properties of both architectures and optimizations performed to mitigate their limitations, a comparative analysis on how both model types perform on different domains, considering; training from initialized weights, transfer learning, self-supervision and inference.

### 4.2.1 Training From Randomly Initialized Weights

Random weight initialization requires the model to learn representations from scratch and; only from that which was fed into the model for training. The performance of transformers trained from scratch and CNNs (He et al., 2015) are compared in a variety of literature (Fauw, 2022; Park & Kim, 2022; Shuang Yu, 2021). On smaller datasets, CNNs seem to have performed better because of the inductive bias inherent within their architecture. However, as the data size increases, transformers are seen to gradually out-perform their CNN counterpart. Training from scratch is seldom practiced in medical imaging due to the small size of most medical data.

### 4.2.2 ImageNet and Same Domain Pre-training

To augment for the size of most domain specific data in computer vision and medical imaging, ImageNet pre-trained weights are often employed for model initialization: this usually results in improved performance (Morid et al., 2021). Researchers have investigated the benefits and extent of transfer learning to ViTs. They discovered that both CNN and ViTs seem to benefit from ImageNet pre-trained weight and same domain pre-training as well, it is also recorded that transformers benefit the most from transfer learning (Hosseinzadeh Taher et al., 2021; Raghu et al., 2019).

### 4.2.3 Self-Supervision

Self-supervision is the most effective solution to the problem of lack of large sized, well-annotated data for modeling in computer vison and medical imaging. The most utilized self-supervised learning schemes; BYOL and DINO, have been found to achieve performance that are comparable to supervised learning schemes. This has prompted their utilization, alongside supervised fine-tuning, in computer vision; achieving state-of-the-art performance (Afouras et al., 2020; Hendrycks et al., 2019; Kolesnikov et al., 2019; Jiaolong Xu et al., 2019). This concept have been attempted on a variety of medical image modalities (R. J. Chen et al., 2022; Jun et al., 2021; Xiyue Wang et al., 2022; B. Zhou et al., 2022) also recording state-of-the-art performances. Additionally, ViTs are found to benefit slightly more from self-supervised pre-training than CNNs.

### 4.2.4 Inference without Fine-Tuning

In cases of extremely scarce data, features extracted from a pre-trained network can be utilized directly for classification and clustering operations. This is most optimal when the pre-trained features are closely related to the target features. One research assessed whether ViTs performed better that CNNs in inference without fine-tuning by applying *k*-NN evaluation on the penultimate layer of a CNN and the CLS token of a ViT. They evaluated for both in-domain and out-domain pre-trained weights and recorded that ViTs performed better in both scenarios (Fauw, 2022).

*Corresponding Author*: Tel; +234-811-940-0230
Email Address: emerald.henry@stu.cu.edu.ng

# REFERENCES

Afouras, T., Owens, A., Chung, J. S., & Zisserman, A. (2020). Self-supervised Learning of Audio-Visual Objects from Video. *Lecture Notes in Computer Science (Including Subseries Lecture Notes in Artificial Intelligence and Lecture Notes in Bioinformatics)*, *12363 LNCS*, 208–224. https://doi.org/10.1007/978-3-030-58523-5_13

Al., K. I. et al. (2022). Whole Slide Image Analysis and Detection of Prostate Cancer using Vision Transformers. *International Conference on Artificial Intelligence in Information and Communication (ICAIIC)*. https://doi.org/doi: 10.1109/ICAIIC54071.2022.9722635.

Almalik, F., Yaqub, M., & Nandakumar, K. (2022). *Self-Ensembling Vision Transformer (SEViT) for Robust Medical Image Classification*. http://arxiv.org/abs/2208.02851

Armato, S. G., McLennan, G., Bidaut, L., McNitt-Gray, M. F., Meyer, C. R., Reeves, A. P., Zhao, B., Aberle, D. R., Henschke, C. I., Hoffman, E. A., Kazerooni, E. A., MacMahon, H., Van Beek, E. J. R., Yankelevitz, D., Biancardi, A. M., Bland, P. H., Brown, M. S., Engelmann, R. M., Laderach, G. E., … Clarke, L. P. (2011). The Lung Image Database Consortium (LIDC) and Image Database Resource Initiative (IDRI): A completed reference database of lung nodules on CT scans. *Medical Physics*, *38*(2), 915–931. https://doi.org/10.1118/1.3528204

Bahdanau, D., Cho, K. H., & Bengio, Y. (2015). Neural machine translation by jointly learning to align and translate. *3rd International Conference on Learning Representations, ICLR 2015 - Conference Track Proceedings*, 1–15.

Bakas, S., Akbari, H., Sotiras, A., Bilello, M., Rozycki, M., Kirby, J. S., Freymann, J. B., Farahani, K., & Davatzikos, C. (2017). Advancing The Cancer Genome Atlas glioma MRI collections with expert segmentation labels and radiomic features. *Scientific Data*, *4*(July), 1–13. https://doi.org/10.1038/sdata.2017.117

Barhoumi, Y., & Ghulam, R. (2021). *Scopeformer: n-CNN-ViT Hybrid Model for Intracranial Hemorrhage Classification*. 2–4. http://arxiv.org/abs/2107.04575

Bhojanapalli, S., Chakrabarti, A., Glasner, D., Li, D., Unterthiner, T., & Veit, A. (2021). Understanding Robustness of Transformers for Image Classification. *Proceedings of the IEEE International Conference on Computer Vision*, 10211–10221. https://doi.org/10.1109/ICCV48922.2021.01007

Campello, V. M., Gkontra, P., Izquierdo, C., Martin-Isla, C., Sojoudi, A., Full, P. M., Maier-Hein, K., Zhang, Y., He, Z., Ma, J., Parreno, M., Albiol, A., Kong, F., Shadden, S. C., Acero, J. C., Sundaresan, V., Saber, M., Elattar, M., Li, H., … Lekadir, K. (2021). Multi-Centre, Multi-Vendor and Multi-Disease Cardiac Segmentation: The MMs Challenge. *IEEE Transactions on Medical Imaging*, *40*(12), 3543–3554. https://doi.org/10.1109/TMI.2021.3090082

Cao, Y., Xu, J., Lin, S., Wei, F., & Hu, H. (2019). GCNet: Non-local networks meet squeeze-excitation networks and beyond. *Proceedings - 2019 International Conference on Computer Vision Workshop, ICCVW 2019*, 1971–1980. https://doi.org/10.1109/ICCVW.2019.00246

Carion, N., Massa, F., Synnaeve, G., Usunier, N., Kirillov, A., & Zagoruyko, S. (2020). End-to-End Object Detection with Transformers. *Lecture Notes in Computer Science (Including Subseries Lecture Notes in Artificial Intelligence and Lecture Notes in Bioinformatics)*, *12346 LNCS*, 213–229. https://doi.org/10.1007/978-3-030-58452-8_13

Caron, M., Touvron, H., Misra, I., Jegou, H., Mairal, J., Bojanowski, P., & Joulin, A. (2021). Emerging Properties in Self-Supervised Vision Transformers. *Proceedings of the IEEE International Conference on Computer Vision*, 9630–9640. https://doi.org/10.1109/ICCV48922.2021.00951

Chen, B., Liu, Y., Zhang, Z., Lu, G., & Zhang, D. (2021). *TransAttUnet: Multi-level Attention-guided U-Net with Transformer for Medical Image Segmentation*. *X*(X), 1–13. http://arxiv.org/abs/2107.05274

Chen, H., Li, C., Wang, G., Li, X., Mamunur Rahaman, M., Sun, H., Hu, W., Li, Y., Liu, W., Sun, C., Ai, S., & Grzegorzek, M. (2022). GasHis-Transformer: A multi-scale visual transformer approach for gastric histopathological image detection. *Pattern Recognition*, *130*(Chen Li), 108827. https://doi.org/10.1016/j.patcog.2022.108827

Chen, J., Frey, E. C., He, Y., Segars, W. P., Li, Y., & Du, Y. (2021). *TransMorph: Transformer for unsupervised medical image registration*. http://arxiv.org/abs/2111.10480

Chen, R. J., Chen, C., Li, Y., Chen, T. Y., Trister, A. D., Krishnan, R. G., & Mahmood, F. (2022). *Scaling Vision Transformers to Gigapixel Images via Hierarchical Self-Supervised Learning*. http://arxiv.org/abs/2206.02647

Chen, R. J., & Krishnan, R. G. (2022). *Self-Supervised Vision Transformers Learn Visual Concepts in Histopathology*. *MIL*, 1–11. http://arxiv.org/abs/2203.00585

Chen, Y., Dai, X., Chen, D., Liu, M., Dong, X., Yuan, L., & Liu, Z. (2021). *Mobile-Former: Bridging MobileNet and Transformer*. http://arxiv.org/abs/2108.05895

Cheng, L., Wei, W., Zhu, F., Liu, Y., & Miao, C. (2021). *Geometry-Entangled Visual Semantic Transformer for Image Captioning*. *14*(8), 1–12. http://arxiv.org/abs/2109.14137

*Corresponding Author*: Tel; +234-811-940-0230
Email Address: emerald.henry@stu.cu.edu.ng

Choi, M., Zhang, Y., Han, K., Wang, X., & Liu, Z. (2022). *Human Eyes Inspired Recurrent Neural Networks are More Robust Against Adversarial Noises*. 1–16. http://arxiv.org/abs/2206.07282

Cordonnier, J.-B., Loukas, A., & Jaggi, M. (2019). *On the Relationship between Self-Attention and Convolutional Layers*. http://arxiv.org/abs/1911.03584

d'Ascoli, S., Touvron, H., Leavitt, M., Morcos, A., Biroli, G., & Sagun, L. (2021). *ConViT: Improving Vision Transformers with Soft Convolutional Inductive Biases*. http://arxiv.org/abs/2103.10697

Dahan, S., Fawaz, A., Williams, L. Z. J., Yang, C., Coalson, T. S., Glasser, M. F., Edwards, A. D., Rueckert, D., & Robinson, E. C. (2022). *Surface Vision Transformers: Attention-Based Modelling applied to Cortical Analysis*. 1–22. http://arxiv.org/abs/2203.16414

Dai, J., Qi, H., Xiong, Y., Li, Y., Zhang, G., Hu, H., & Wei, Y. (2017). Deformable Convolutional Networks. *Proceedings of the IEEE International Conference on Computer Vision*, *2017-Octob*, 764–773. https://doi.org/10.1109/ICCV.2017.89

Dai, Y., Gao, Y., & Liu, F. (2021). Transmed: Transformers advance multi-modal medical image classification. *Diagnostics*, *11*(8), 1–8. https://doi.org/10.3390/diagnostics11081384

Dai, Z., Liu, H., Le, Q. V., & Tan, M. (2021). CoAtNet: Marrying Convolution and Attention for All Data Sizes. *Advances in Neural Information Processing Systems*, *5*, 3965–3977.

Darby, M. J., Barron, D. A., & Hyland, R. E. (2012). *Oxford Handbook of Medical Edited by*.

Devlin, J., Chang, M. W., Lee, K., & Toutanova, K. (2019). BERT: Pre-training of deep bidirectional transformers for language understanding. *NAACL HLT 2019 - 2019 Conference of the North American Chapter of the Association for Computational Linguistics: Human Language Technologies - Proceedings of the Conference*, *1*(Mlm), 4171–4186.

Dosovitskiy, A., Beyer, L., Kolesnikov, A., Weissenborn, D., Zhai, X., Unterthiner, T., Dehghani, M., Minderer, M., Heigold, G., Gelly, S., Uszkoreit, J., & Houlsby, N. (2020). *An Image is Worth 16x16 Words: Transformers for Image Recognition at Scale*. http://arxiv.org/abs/2010.11929

Fauw. (2022). *Can Transformers Replace Cnns for Medical Image Classification?* 1–15.

Feng, C.-M., Yan, Y., Chen, G., Xu, Y., Hu, Y., Shao, L., & Fu, H. (2022). Multi-Modal Transformer for Accelerated MR Imaging. *IEEE Transactions on Medical Imaging*, *XX*(Xx), 1–1. https://doi.org/10.1109/tmi.2022.3180228

Feng, C. M., Yan, Y., Fu, H., Chen, L., & Xu, Y. (2021). Task Transformer Network for Joint MRI Reconstruction and Super-Resolution. *Lecture Notes in Computer Science (Including Subseries Lecture Notes in Artificial Intelligence and Lecture Notes in Bioinformatics)*, *12906 LNCS*, 307–317. https://doi.org/10.1007/978-3-030-87231-1_30

Gao., X. (2021). COVID-VIT: Classification of Covid-19 from CT chest images based on vision transformer models. *1999*(December), 1–6.

Gao, Y., Zhou, M., Liu, D., Yan, Z., Zhang, S., & Metaxas, D. N. (2022). *A Data-scalable Transformer for Medical Image Segmentation: Architecture, Model Efficiency, and Benchmark*. Cv, 1–39. http://arxiv.org/abs/2203.00131

Gao, Z., Hong, B., Zhang, X., Li, Y., Jia, C., Wu, J., Wang, C., Meng, D., & Li, C. (2021). Instance-Based Vision Transformer for Subtyping of Papillary Renal Cell Carcinoma in Histopathological Image. *Lecture Notes in Computer Science (Including Subseries Lecture Notes in Artificial Intelligence and Lecture Notes in Bioinformatics)*, *12908 LNCS*, 299–308. https://doi.org/10.1007/978-3-030-87237-3_29

Ghaderzadeh, M., & Asadi, F. (2021). Deep Learning in the Detection and Diagnosis of COVID-19 Using Radiology Modalities: A Systematic Review. *Journal of Healthcare Engineering*, *2021*. https://doi.org/10.1155/2021/6677314

Graves, A., Mohamed, A. R., & Hinton, G. (2013). Speech recognition with deep recurrent neural networks. *ICASSP, IEEE International Conference on Acoustics, Speech and Signal Processing - Proceedings*, *3*, 6645–6649. https://doi.org/10.1109/ICASSP.2013.6638947

Gunraj, H., Sabri, A., Koff, D., & Wong, A. (2022). COVID-Net CT-2: Enhanced Deep Neural Networks for Detection of COVID-19 From Chest CT Images Through Bigger, More Diverse Learning. *Frontiers in Medicine*, *8*, 1–15. https://doi.org/10.3389/fmed.2021.729287

Haghanifar, A., Majdabadi, M. M., Choi, Y., Deivalakshmi, S., & Ko, S. (2022). COVID-CXNet: Detecting COVID-19 in frontal chest X-ray images using deep learning. *Multimedia Tools and Applications*. https://doi.org/10.1007/s11042-022-12156-z

Hajeb Mohammad Alipour, S., Rabbani, H., & Akhlaghi, M. R. (2012). Diabetic retinopathy grading by digital curvelet transform. *Computational and Mathematical Methods in Medicine*, *2012*. https://doi.org/10.1155/2012/761901

Hatamizadeh, A., Tang, Y., Nath, V., Yang, D., Myronenko, A., Landman, B., Roth, H. R., & Xu, D. (2022). UNETR: Transformers for 3D Medical Image Segmentation. *Proceedings - 2022 IEEE/CVF Winter Conference on*

*Corresponding Author*: Tel; +234-811-940-0230
Email Address: emerald.henry@stu.cu.edu.ng

Applications of Computer Vision, WACV 2022, 1748–1758. https://doi.org/10.1109/WACV51458.2022.00181

Hatamizadeh, A., Xu, Z., Yang, D., Li, W., Roth, H., & Xu, D. (2022). *UNetFormer: A Unified Vision Transformer Model and Pre-Training Framework for 3D Medical Image Segmentation*. 1–12. http://arxiv.org/abs/2204.00631

He, K., Zhang, X., Ren, S., & Sun, J. (2015). Delving deep into rectifiers: Surpassing human-level performance on imagenet classification. *Proceedings of the IEEE International Conference on Computer Vision*, *2015 Inter*, 1026–1034. https://doi.org/10.1109/ICCV.2015.123

He, K., Zhang, X., Ren, S., & Sun, J. (2016). Deep residual learning for image recognition. *Proceedings of the IEEE Computer Society Conference on Computer Vision and Pattern Recognition*, *2016-Decem*, 770–778. https://doi.org/10.1109/CVPR.2016.90

He, S., Grant, P. E., & Ou, Y. (2022). Global-Local Transformer for Brain Age Estimation. *IEEE Transactions on Medical Imaging*, *41*(1), 213–224. https://doi.org/10.1109/TMI.2021.3108910

Heidari, M., Kazerouni, A., Soltany, M., Azad, R., Aghdam, E. K., Cohen-Adad, J., & Merhof, D. (2022). *HiFormer: Hierarchical Multi-scale Representations Using Transformers for Medical Image Segmentation*. 1–11. http://arxiv.org/abs/2207.08518

Heller, N., Isensee, F., Maier-hein, K. H., Hou, X., Xie, C., Li, F., Nan, Y., Mu, G., Lin, Z., Han, M., Yao, G., Zhang, Y., Wang, Y., Hou, F., Yang, J., Xiong, G., Tian, J., Zhong, C., Rickman, J., … States, U. (2022). *HHS Public Access*. 1–39. https://doi.org/10.1016/j.media.2020.101821.The

Hendrycks, D., Mazeika, M., Kadavath, S., & Song, D. (2019). Using self-supervised learning can improve model robustness and uncertainty. *Advances in Neural Information Processing Systems*, *32*(NeurIPS).

Hoppe, S., & Toussaint, M. (2020). *Qgraph-bounded Q-learning: Stabilizing Model-Free Off-Policy Deep Reinforcement Learning*. http://arxiv.org/abs/2007.07582

Hosseinzadeh Taher, M. R., Haghighi, F., Feng, R., Gotway, M. B., & Liang, J. (2021). A Systematic Benchmarking Analysis of Transfer Learning for Medical Image Analysis. *Lecture Notes in Computer Science (Including Subseries Lecture Notes in Artificial Intelligence and Lecture*

*Notes in Bioinformatics)*, *12968 LNCS*, 3–13. https://doi.org/10.1007/978-3-030-87722-4_1

Hsu, C.-C., Chen, G.-L., & Wu, M.-H. (2021). *Visual Transformer with Statistical Test for COVID-19 Classification*. *2*(1). http://arxiv.org/abs/2107.05334

Hu, J., Shen, L., Albanie, S., Sun, G., & Wu, E. (2020). Squeeze-and-Excitation Networks. *IEEE Transactions on Pattern Analysis and Machine Intelligence*, *42*(8), 2011–2023. https://doi.org/10.1109/TPAMI.2019.2913372

Huang, S., Li, J., Xiao, Y., Shen, N., & Xu, T. (2022). RTNet: Relation Transformer Network for Diabetic Retinopathy Multi-Lesion Segmentation. *IEEE Transactions on Medical Imaging*, *41*(6), 1596–1607. https://doi.org/10.1109/TMI.2022.3143833

Huang, Z., Wang, X., Huang, L., Huang, C., Wei, Y., & Liu, W. (2019). CCNet: Criss-cross attention for semantic segmentation. *Proceedings of the IEEE International Conference on Computer Vision*, *2019-Octob*(July), 603–612. https://doi.org/10.1109/ICCV.2019.00069

Jaeger, S., Candemir, S., Antani, S., Wáng, Y.-X. J., Lu, P.-X., & Thoma, G. (2014). Two public chest X-ray datasets for computer-aided screening of pulmonary diseases. *Quantitative Imaging in Medicine and Surgery*, *4*(6), 475–477. https://doi.org/10.3978/j.issn.2223-4292.2014.11.20

Jang, J., & Hwang, D. (2022). *M3T: three-dimensional Medical image classifier using Multi-plane and Multi-slice Transformer*. 20718–20729.

Jiang, Y., Chang, S., & Wang, Z. (2021). TransGAN: Two Pure Transformers Can Make One Strong GAN, and That Can Scale Up. *Advances in Neural Information Processing Systems*, *18*(NeurIPS), 14745–14758.

Jun, E., Jeong, S., Heo, D.-W., & Suk, H.-I. (2021). *Medical Transformer: Universal Brain Encoder for 3D MRI Analysis*. 1–9. http://arxiv.org/abs/2104.13633

Kather, J. N., Zöllner, F. G., Bianconi, F., Melchers, S. M., Schad, L. R., Gaiser, T., Marx, A., & Weis, C.-A. (2016). *Collection of textures in colorectal cancer histology*. https://doi.org/10.5281/ZENODO.53169

Kavur, A. E., Gezer, N. S., Barış, M., Aslan, S., Conze, P. H., Groza, V., Pham, D. D., Chatterjee, S., Ernst, P., Özkan, S., Baydar, B., Lachinov, D., Han, S., Pauli, J., Isensee, F., Perkonigg, M., Sathish, R., Rajan, R., Sheet, D., … Selver, M. A. (2021). CHAOS Challenge - combined (CT-MR) healthy abdominal organ segmentation. *Medical Image Analysis*, *69*. https://doi.org/10.1016/j.media.2020.101950

Kermany, Daniel; Zhang, Kang; Goldbaum, M. (2018). (n.d.). Large Dataset of Labeled Optical Coherence Tomography (OCT) and Chest X-Ray Images. *Mendeley Data, V3, Doi: 10.17632/Rscbjbr9sj.3*.

Kermany, D. S., Goldbaum, M., Cai, W., Valentim, C. C. S., Liang, H., Baxter, S. L., McKeown, A., Yang, G., Wu, X., Yan, F., Dong, J., Prasadha, M. K., Pei, J., Ting, M., Zhu,

*Corresponding Author*: Tel; +234-811-940-0230
Email Address: emerald.henry@stu.cu.edu.ng

J., Li, C., Hewett, S., Dong, J., Ziyar, I., … Zhang, K. (2018). Identifying Medical Diagnoses and Treatable Diseases by Image-Based Deep Learning. *Cell*, *172*(5), 1122-1131.e9. https://doi.org/10.1016/j.cell.2018.02.010

Kim, B. J., Choi, H., Jang, H., Lee, D. G., Jeong, W., & Kim, S. W. (2021). *Dead Pixel Test Using Effective Receptive Field*. http://arxiv.org/abs/2108.13576

Kolesnikov, A., Zhai, X., & Beyer, L. (2019). Revisiting self-supervised visual representation learning. *Proceedings of the IEEE Computer Society Conference on Computer Vision and Pattern Recognition*, *2019-June*, 1920–1929. https://doi.org/10.1109/CVPR.2019.00202

Kollias, D., Arsenos, A., Soukissian, L., & Kollias, S. (2021). MIA-COV19D: COVID-19 Detection through 3-D Chest CT Image Analysis. *Proceedings of the IEEE International Conference on Computer Vision*, *2021-Octob*, 537–544. https://doi.org/10.1109/ICCVW54120.2021.00066

Lan, Z., Chen, M., Goodman, S., Gimpel, K., Sharma, P., & Soricut, R. (2019). *ALBERT: A Lite BERT for Self-supervised Learning of Language Representations*. 1–17. http://arxiv.org/abs/1909.11942

Li, C., Chen, H., Li, X., Xu, N., Hu, Z., Xue, D., Qi, S., Ma, H., Zhang, L., & Sun, H. (2020). A review for cervical histopathology image analysis using machine vision approaches. In *Artificial Intelligence Review* (Vol. 53, Issue 7). https://doi.org/10.1007/s10462-020-09808-7

Li, Jiangyun, Wang, W., Chen, C., Zhang, T., Zha, S., Wang, J., & Yu, H. (2022). *TransBTSV2: Towards Better and More Efficient Volumetric Segmentation of Medical Images*. 1–13. http://arxiv.org/abs/2201.12785

Li, Jingxing, Yang, Z., & Yu, Y. (2021). A Medical AI Diagnosis Platform Based on Vision Transformer for Coronavirus. *2021 IEEE International Conference on Computer Science, Electronic Information Engineering and Intelligent Control Technology, CEI 2021*, 246–252. https://doi.org/10.1109/CEI52496.2021.9574576

Li, Jun, Chen, J., Tang, Y., Wang, C., Landman, B. A., & Zhou, S. K. (2022). *Transforming medical imaging with Transformers? A comparative review of key properties, current progresses, and future perspectives*. http://arxiv.org/abs/2206.01136

Li, N., Liu, S., Liu, Y., Zhao, S., & Liu, M. (2019). Neural speech synthesis with transformer network. *33rd AAAI Conference on Artificial Intelligence, AAAI 2019, 31st Innovative Applications of Artificial Intelligence Conference, IAAI 2019 and the 9th AAAI Symposium on Educational Advances in Artificial Intelligence, EAAI 2019*, 6706–6713. https://doi.org/10.1609/aaai.v33i01.33016706

Li, S., Sui, X., Luo, X., Xu, X., Liu, Y., & Goh, R. (2021). Medical Image Segmentation Using Squeeze-and-Expansion Transformers. *IJCAI International Joint Conference on Artificial Intelligence*, 807–815. https://doi.org/10.24963/ijcai.2021/112

Li, Yanghao, Mao, H., Girshick, R., & He, K. (2022). *Exploring Plain Vision Transformer Backbones for Object Detection*. 1–21. http://arxiv.org/abs/2203.16527

Li, Yehao, Yao, T., Pan, Y., & Mei, T. (2022). Contextual Transformer Networks for Visual Recognition. *IEEE Transactions on Pattern Analysis and Machine Intelligence*. https://doi.org/10.1109/TPAMI.2022.3164083

Li, Yunxiang, Wang, S., Wang, J., Zeng, G., Liu, W., Zhang, Q., Jin, Q., & Wang, Y. (2021). GT U-Net: A U-Net Like Group Transformer Network for Tooth Root Segmentation. *Lecture Notes in Computer Science (Including Subseries Lecture Notes in Artificial Intelligence and Lecture Notes in Bioinformatics)*, *12966 LNCS*, 386–395. https://doi.org/10.1007/978-3-030-87589-3_40

Liang, S., Zhang, W., & Gu, Y. (2021). A hybrid and fast deep learning framework for Covid-19 detection via 3D Chest CT Images. *Proceedings of the IEEE International Conference on Computer Vision*, *2021-Octob*, 508–512. https://doi.org/10.1109/ICCVW54120.2021.00062

Liu, Y., Ott, M., Goyal, N., Du, J., Joshi, M., Chen, D., Levy, O., Lewis, M., Zettlemoyer, L., & Stoyanov, V. (2019). *RoBERTa: A Robustly Optimized BERT Pretraining Approach*. *1*. http://arxiv.org/abs/1907.11692

Liu, Ze, Lin, Y., Cao, Y., Hu, H., Wei, Y., Zhang, Z., Lin, S., & Guo, B. (2021). Swin Transformer: Hierarchical Vision Transformer using Shifted Windows. *Proceedings of the IEEE International Conference on Computer Vision*, 9992–10002. https://doi.org/10.1109/ICCV48922.2021.00986

Liu, Zhuang, Mao, H., Wu, C.-Y., Feichtenhofer, C., Darrell, T., & Xie, S. (2022). *A ConvNet for the 2020s*. http://arxiv.org/abs/2201.03545

Louis, M. (2013). 20:21. *Canadian Journal of Emergency Medicine*, *15*(3), 190. https://doi.org/10.2310/8000.2013.131108

Luthra, A., Sulakhe, H., Mittal, T., Iyer, A., & Yadav, S. (2021). *Eformer: Edge Enhancement based Transformer for Medical Image Denoising*. http://arxiv.org/abs/2109.08044

Maguolo, G., & Nanni, L. (2019). A Critic Evaluation of Methods for COVID-19. *Unipd*, 1–10.

Matsoukas, C., Haslum, J. F., Söderberg, M., & Smith, K. (2021). *Is it Time to Replace CNNs with Transformers for*

*Corresponding Author*: Tel; +234-811-940-0230
Email Address: emerald.henry@stu.cu.edu.ng

*Medical Images?* http://arxiv.org/abs/2108.09038

Mendrik, A. M., Vincken, K. L., Kuijf, H. J., Breeuwer, M., Bouvy, W. H., De Bresser, J., Alansary, A., De Bruijne, M., Carass, A., El-Baz, A., Jog, A., Katyal, R., Khan, A. R., Van Der Lijn, F., Mahmood, Q., Mukherjee, R., Van Opbroek, A., Paneri, S., Pereira, S., … Viergever, M. A. (2015). MRBrainS Challenge: Online Evaluation Framework for Brain Image Segmentation in 3T MRI Scans. *Computational Intelligence and Neuroscience*, *2015*. https://doi.org/10.1155/2015/813696

Mondal, A. K., Bhattacharjee, A., Singla, P., & Prathosh, A. P. (2022). XViTCOS: Explainable Vision Transformer Based COVID-19 Screening Using Radiography. *IEEE Journal of Translational Engineering in Health and Medicine*, *10*(December 2021). https://doi.org/10.1109/JTEHM.2021.3134096

Moreira, I. C., Amaral, I., Domingues, I., Cardoso, A., Cardoso, M. J., & Cardoso, J. S. (2012). INbreast: Toward a Full-field Digital Mammographic Database. *Academic Radiology*, *19*(2), 236–248. https://doi.org/10.1016/j.acra.2011.09.014

Morid, M. A., Borjali, A., & Del Fiol, G. (2021). A scoping review of transfer learning research on medical image analysis using ImageNet. *Computers in Biology and Medicine*, *128*(408). https://doi.org/10.1016/j.compbiomed.2020.104115

Pan, J., Zhang, H., Wu, W., Gao, Z., & Wu, W. (2022). Multi-domain integrative Swin transformer network for sparse-view tomographic reconstruction. *Patterns*, *3*(6). https://doi.org/10.1016/j.patter.2022.100498

Park, N., & Kim, S. (2022). *How Do Vision Transformers Work?* http://arxiv.org/abs/2202.06709

Park, S., Kim, G., Kim, J., Kim, B., & Ye, J. C. (2021). Federated Split Vision Transformer for COVID-19 CXR Diagnosis using Task-Agnostic Training. *Advances in Neural Information Processing Systems*, *29*(NeurIPS), 24617–24630.

Park, S., Kim, G., Oh, Y., Seo, J. B., Lee, S. M., Lim, J., Park, C. M., Ye, J. C., & Kim, J. H. (2022). *diagnosis through knowledge distillation*. 1–11. https://doi.org/10.1038/s41467-022-31514-x

Park, S., & Ye, J. C. (2022). *Multi-Task Distributed Learning using Vision Transformer with Random Patch Permutation*. 1–10. http://arxiv.org/abs/2204.03500

Parvaiz, A., Khalid, M. A., Zafar, R., Ameer, H., Ali, M., & Fraz, M. M. (2022). *Vision Transformers in Medical Computer Vision -- A Contemplative Retrospection*. 0–3. http://arxiv.org/abs/2203.15269

Peng, C., Myronenko, A., Hatamizadeh, A., Nath, V., Siddiquee, M. M. R., He, Y., Xu, D., Chellappa, R., &

*Corresponding Author*: Tel; +234-811-940-0230
Email Address: emerald.henry@stu.cu.edu.ng

Yang, D. (2021). *HyperSegNAS: Bridging One-Shot Neural Architecture Search with 3D Medical Image Segmentation using HyperNet*. http://arxiv.org/abs/2112.10652

Peng, Z., Huang, W., Gu, S., Xie, L., Wang, Y., Jiao, J., & Ye, Q. (2021). Conformer: Local Features Coupling Global Representations for Visual Recognition. *Proceedings of the IEEE International Conference on Computer Vision*, 357–366. https://doi.org/10.1109/ICCV48922.2021.00042

Peterfy, C. G., Schneider, E., & Nevitt, M. (2008). The osteoarthritis initiative: report on the design rationale for the magnetic resonance imaging protocol for the knee. *Osteoarthritis and Cartilage*, *16*(12), 1433–1441. https://doi.org/10.1016/j.joca.2008.06.016

Radford, A., Narasimhan, K., Salimans, T., & Sutskever, I. (2018). Improving Language Understanding by Generative Pre-Training. *Homology, Homotopy and Applications*.

Raghu, M., Unterthiner, T., Kornblith, S., Zhang, C., & Dosovitskiy, A. (2021). Do Vision Transformers See Like Convolutional Neural Networks? *Advances in Neural Information Processing Systems*, *15*(NeurIPS), 12116–12128.

Raghu, M., Zhang, C., Kleinberg, J., & Bengio, S. (2019). Transfusion: Understanding transfer learning for medical imaging. *Advances in Neural Information Processing Systems*, *32*(NeurIPS).

Rahimzadeh, M., Attar, A., & Mohammad, S. (2020). *Since January 2020 Elsevier has created a COVID-19 resource centre with free information in English and Mandarin on the novel coronavirus COVID- 19 . The COVID-19 resource centre is hosted on Elsevier Connect , the company ' s public news and information . January*.

Ramachandran, P., Bello, I., Parmar, N., Levskaya, A., Vaswani, A., & Shlens, J. (2019). Stand-alone self-attention in vision models. *Advances in Neural Information Processing Systems*, *32*.

Rui Sun, Y. L. (2021). *Lesion-Aware Transformers for Diabetic Retinopathy Grading*.

Saeed, F., Hussain, M., Member, S., IEEE, Aboalsamh, H. A., Member, S., IEEE, Adel, F. Al, & Owaifeer, A. M. Al. (2021). *Diabetic Retinopathy Screening Using Custom-Designed Convolutional Neural Network*. *i*. http://arxiv.org/abs/2110.03877

Sak, H., Senior, A., & Beaufays, F. (2014). *Long Short-Term Memory Based Recurrent Neural Network Architectures for Large Vocabulary Speech Recognition*. *Cd*. http://arxiv.org/abs/1402.1128

Setio, A. A. A., Traverso, A., de Bel, T., Berens, M. S. N.,

Bogaard, C. van den, Cerello, P., Chen, H., Dou, Q., Fantacci, M. E., Geurts, B., Gugten, R. van der, Heng, P. A., Jansen, B., de Kaste, M. M. J., Kotov, V., Lin, J. Y. H., Manders, J. T. M. C., Sóñora-Mengana, A., García-Naranjo, J. C., … Xie, P. (2017). Validation, comparison, and combination of algorithms for automatic detection of pulmonary nodules in computed tomography images: The LUNA16 challenge. *Medical Image Analysis*, *42*, 1–13. https://doi.org/10.1016/j.media.2017.06.015

Shao, Z., Bian, H., Chen, Y., Wang, Y., Zhang, J., Ji, X., & Zhang, Y. (2021). TransMIL: Transformer based Correlated Multiple Instance Learning for Whole Slide Image Classification. *Advances in Neural Information Processing Systems*, *3*(NeurIPS), 2136–2147.

Shen, L., Zheng, J., Lee, E. H., Shpanskaya, K., McKenna, E. S., Atluri, M. G., Plasto, D., Mitchell, C., Lai, L. M., Guimaraes, C. V., Dahmoush, H., Chueh, J., Halabi, S. S., Pauly, J. M., Xing, L., Lu, Q., Oztekin, O., Kline-Fath, B. M., & Yeom, K. W. (2022). Attention-guided deep learning for gestational age prediction using fetal brain MRI. *Scientific Reports*, *12*(1), 1–10. https://doi.org/10.1038/s41598-022-05468-5

Sheng Wang, Z. Z. (2021). *3DMeT: 3D Medical Image Transformer for Knee Cartilage Defect Assessment*.

Shiraishi, J., Katsuragawa, S., Ikezoe, J., Matsumoto, T., Kobayashi, T., Komatsu, K. I., Matsui, M., Fujita, H., Kodera, Y., & Doi, K. (2000). Development of a digital image database for chest radiographs with and without a lung nodule: Receiver operating characteristic analysis of radiologists' detection of pulmonary nodules. *American Journal of Roentgenology*, *174*(1), 71–74. https://doi.org/10.2214/ajr.174.1.1740071

Shuang Yu, K. M. (2021). *MIL-VT: Multiple Instance Learning Enhanced Vision Transformer for Fundus Image Classification*.

Soares, E., & Angelov, P. (2020). A large dataset of real patients CT scans for COVID-19 identification. *Harv. Dataverse*, *1*, 1–8.

Sørensen, L., Shaker, S. B., & De Bruijne, M. (2010). Quantitative analysis of pulmonary emphysema using local binary patterns. *IEEE Transactions on Medical Imaging*, *29*(2), 559–569. https://doi.org/10.1109/TMI.2009.2038575

Srinivas, A., Lin, T. Y., Parmar, N., Shlens, J., Abbeel, P., & Vaswani, A. (2021). Bottleneck transformers for visual recognition. *Proceedings of the IEEE Computer Society Conference on Computer Vision and Pattern Recognition*, *Figure 1*, 16514–16524. https://doi.org/10.1109/CVPR46437.2021.01625

Stirrat, C. G., Alam, S. R., MacGillivray, T. J., Gray, C. D., Dweck, M. R., Raftis, J., Jenkins, W. S. A., Wallace, W. A., Pessotto, R., Lim, K. H. H., Mirsadraee, S., Henriksen, P. A., Semple, S. I. K., & Newby, D. E. (2017). Ferumoxytol-enhanced magnetic resonance imaging assessing inflammation after myocardial infarction. *Heart*, *103*(19), 1528–1535. https://doi.org/10.1136/heartjnl-2016-311018

Sudlow, C., Gallacher, J., Allen, N., Beral, V., Burton, P., Danesh, J., Downey, P., Elliott, P., Green, J., Landray, M., Liu, B., Matthews, P., Ong, G., Pell, J., Silman, A., Young, A., Sprosen, T., Peakman, T., & Collins, R. (2015). UK Biobank: An Open Access Resource for Identifying the Causes of a Wide Range of Complex Diseases of Middle and Old Age. *PLoS Medicine*, *12*(3), 1–10. https://doi.org/10.1371/journal.pmed.1001779

Sun, R., & Pang, Y. (n.d.). *Efficient Lung Cancer Image Classification and Segmentation Algorithm Based on Improved Swin Transformer*. 1–15.

Susanti, H. D., Arfamaini, R., Sylvia, M., Vianne, A., D, Y. H., D, H. L., Muslimah, M. muslimah, Saletti-cuesta, L., Abraham, C., Sheeran, P., Adiyoso, W., Wilopo, W., Brossard, D., Wood, W., Cialdini, R., Groves, R. M., Chan, D. K. C., Zhang, C. Q., Josefsson, K. W., … Aryanta, I. R. (2017).

Tabik, S., Gomez-Rios, A., Martin-Rodriguez, J. L., Sevillano-Garcia, I., Rey-Area, M., Charte, D., Guirado, E., Suarez, J. L., Luengo, J., Valero-Gonzalez, M. A., Garcia-Villanova, P., Olmedo-Sanchez, E., & Herrera, F. (2020). COVIDGR Dataset and COVID-SDNet Methodology for Predicting COVID-19 Based on Chest X-Ray Images. *IEEE Journal of Biomedical and Health Informatics*, *24*(12), 3595–3605. https://doi.org/10.1109/JBHI.2020.3037127

Tan, M., & Le, Q. V. (2019). EfficientNet: Rethinking model scaling for convolutional neural networks. *36th International Conference on Machine Learning, ICML 2019*, *2019-June*, 10691–10700.

Tang, Y., Yang, D., Li, W., Roth, H., Landman, B., Xu, D., Nath, V., & Hatamizadeh, A. (2021). *Self-Supervised Pre-Training of Swin Transformers for 3D Medical Image Analysis*. 20730–20740. http://arxiv.org/abs/2111.14791

Topal, M. O., Bas, A., & van Heerden, I. (2021). *Exploring Transformers in Natural Language Generation: GPT, BERT, and XLNet*. http://arxiv.org/abs/2102.08036

Touvron, H., Cord, M., Douze, M., Massa, F., Sablayrolles, A., & Jégou, H. (2020). *Training data-efficient image transformers & distillation through attention*. 1–22. http://arxiv.org/abs/2012.12877

Tragakis, A., Kaul, C., Murray-Smith, R., & Husmeier, D. (2022). *The Fully Convolutional Transformer for Medical Image Segmentation*. 1–16.

*Corresponding Author*: Tel; +234-811-940-0230
Email Address: emerald.henry@stu.cu.edu.ng

http://arxiv.org/abs/2206.00566

Tummala, S. (2022). *Brain Tumor Classification from MRI using Vision Transformers Ensembling*. 1–17.

Vaswani, A., Ramachandran, P., Srinivas, A., Parmar, N., Hechtman, B., & Shlens, J. (2021). Scaling local self-attention for parameter efficient visual backbones. *Proceedings of the IEEE Computer Society Conference on Computer Vision and Pattern Recognition*, 12889–12899. https://doi.org/10.1109/CVPR46437.2021.01270

Vaswani, A., Shazeer, N., Parmar, N., Uszkoreit, J., Jones, L., Gomez, A. N., Kaiser, Ł., & Polosukhin, I. (2017). Attention is all you need. *Advances in Neural Information Processing Systems*, 2017-Decem(Nips), 5999–6009.

Vayá, M. de la I., Saborit, J. M., Montell, J. A., Pertusa, A., Bustos, A., Cazorla, M., Galant, J., Barber, X., Orozco-Beltrán, D., García-García, F., Caparrós, M., González, G., & Salinas, J. M. (2020). *BIMCV COVID-19+: a large annotated dataset of RX and CT images from COVID-19 patients*. 1–22. http://arxiv.org/abs/2006.01174

Wang, D., Wu, Z., & Yu, H. (2021). TED-Net: Convolution-Free T2T Vision Transformer-Based Encoder-Decoder Dilation Network for Low-Dose CT Denoising. *Lecture Notes in Computer Science (Including Subseries Lecture Notes in Artificial Intelligence and Lecture Notes in Bioinformatics)*, 12966 LNCS, 416–425. https://doi.org/10.1007/978-3-030-87589-3_43

Wang, Li, Nie, D., Li, G., Puybareau, É., Dolz, J., Zhang, Q., Wang, F., Xia, J., Wu, Z., Chen, J. W., Thung, K. H., Bui, T. D., Shin, J., Zeng, G., Zheng, G., Fonov, V. S., Doyle, A., Xu, Y., Moeskops, P., … Shen, D. (2019). Benchmark on automatic six-month-old infant brain segmentation algorithms: The iSeg-2017 challenge. *IEEE Transactions on Medical Imaging*, 38(9), 2219–2230. https://doi.org/10.1109/TMI.2019.2901712

Wang, Linda, Lin, Z. Q., & Wong, A. (2020). COVID-Net: a tailored deep convolutional neural network design for detection of COVID-19 cases from chest X-ray images. *Scientific Reports*, 10(1), 1–12. https://doi.org/10.1038/s41598-020-76550-z

Wang, Q., Wu, B., Zhu, P., Li, P., Zuo, W., & Hu, Q. (2020). ECA-Net: Efficient channel attention for deep convolutional neural networks. *Proceedings of the IEEE Computer Society Conference on Computer Vision and Pattern Recognition*, 11531–11539. https://doi.org/10.1109/CVPR42600.2020.01155

Wang, W., Xie, E., Li, X., Fan, D.-P., Song, K., Liang, D., Lu, T., Luo, P., & Shao, L. (2021). *PVT_Wang_ICCV_2021*. 568–578. http://arxiv.org/abs/2102.12122

Wang, Xiaolong, Girshick, R., Gupta, A., & He, K. (2018). Non-local Neural Networks. *Proceedings of the IEEE Computer Society Conference on Computer Vision and Pattern Recognition*, 7794–7803. https://doi.org/10.1109/CVPR.2018.00813

Wang, Xiyue, Yang, S., Zhang, J., Wang, M., Zhang, J., & Yang, W. (2022). Transformer-based unsupervised contrastive learning for histopathological image classification. *Medical Image Analysis*, 81(June), 102559. https://doi.org/10.1016/j.media.2022.102559

Weinstein, J. N., Collisson, E. A., Mills, G. B., Shaw, K. R. M., Ozenberger, B. A., Ellrott, K., Sander, C., Stuart, J. M., Chang, K., Creighton, C. J., Davis, C., Donehower, L., Drummond, J., Wheeler, D., Ally, A., Balasundaram, M., Birol, I., Butterfield, Y. S. N., Chu, A., … Kling, T. (2013). The cancer genome atlas pan-cancer analysis project. *Nature Genetics*, 45(10), 1113–1120. https://doi.org/10.1038/ng.2764

Weng, W., & Zhu, X. (2021). INet: Convolutional Networks for Biomedical Image Segmentation. *IEEE Access*, 9, 16591–16603. https://doi.org/10.1109/ACCESS.2021.3053408

Winata, G. I., Madotto, A., Lin, Z., Liu, R., Yosinski, J., & Fung, P. (2021). *Language Models are Few-shot Multilingual Learners*. 1–15. https://doi.org/10.18653/v1/2021.mrl-1.1

Woo, S., Park, J., Lee, J. Y., & Kweon, I. S. (2018). CBAM: Convolutional block attention module. *Lecture Notes in Computer Science (Including Subseries Lecture Notes in Artificial Intelligence and Lecture Notes in Bioinformatics)*, 11211 LNCS, 3–19. https://doi.org/10.1007/978-3-030-01234-2_1

Wu, H., Xiao, B., Codella, N., Liu, M., Dai, X., Yuan, L., & Zhang, L. (2021). CvT: Introducing Convolutions to Vision Transformers. *Proceedings of the IEEE International Conference on Computer Vision*, 22–31. https://doi.org/10.1109/ICCV48922.2021.00009

Xia, Y., Yao, J., Lu, L., Huang, L., Xie, G., Xiao, J., Yuille, A., Cao, K., & Zhang, L. (2021). Effective Pancreatic Cancer Screening on Non-contrast CT Scans via Anatomy-Aware Transformers. *Lecture Notes in Computer Science (Including Subseries Lecture Notes in Artificial Intelligence and Lecture Notes in Bioinformatics)*, 12905 LNCS, 259–269. https://doi.org/10.1007/978-3-030-87240-3_25

Xia, Z., Pan, X., Song, S., Li, L. E., & Huang, G. (2022). *Vision Transformer with Deformable Attention*. http://arxiv.org/abs/2201.00520

Xie, Y., Zhang, J., Shen, C., & Xia, Y. (2021). CoTr: Efficiently Bridging CNN and Transformer for 3D Medical Image Segmentation. *Lecture Notes in Computer Science (Including Subseries Lecture Notes in Artificial*

*Corresponding Author*: Tel; +234-811-940-0230
Email Address: emerald.henry@stu.cu.edu.ng

*Intelligence and Lecture Notes in Bioinformatics)*, *12903 LNCS*, 171–180. https://doi.org/10.1007/978-3-030-87199-4_16

Xie, Y., Zhang, J., Xia, Y., & Wu, Q. (2021). *Unified 2D and 3D Pre-training for Medical Image classification and Segmentation*. http://arxiv.org/abs/2112.09356

Xu, H., Su, X., & Wang, D. (2022). CNN-based Local Vision Transformer for COVID-19 Diagnosis. In *Woodstock '18: ACM Symposium on Neural Gaze Detection, June 03â•fi05, 2018, Woodstock, NY* (Vol. 1, Issue 1). Association for Computing Machinery. http://arxiv.org/abs/2207.02027

Xu, Jiaolong, Xiao, L., & Lopez, A. M. (2019). Self-supervised domain adaptation for computer vision tasks. *IEEE Access*, *7*, 156694–156706. https://doi.org/10.1109/ACCESS.2019.2949697

Xu, Junshen, Moyer, D., Grant, P. E., Golland, P., Iglesias, J. E., & Adalsteinsson, E. (2022). *SVoRT: Iterative Transformer for Slice-to-Volume Registration in Fetal Brain MRI*. http://arxiv.org/abs/2206.10802

Xu, K., Ba, J. L., Kiros, R., Cho, K., Courville, A., Salakhutdinov, R., Zemel, R. S., & Bengio, Y. (2015). Show, attend and tell: Neural image caption generation with visual attention. *32nd International Conference on Machine Learning, ICML 2015*, *3*, 2048–2057.

Yang, X., He, X., Zhao, J., Zhang, Y., Zhang, S., & Xie, P. (2020). *COVID-CT-Dataset: A CT Scan Dataset about COVID-19*. 1–14. http://arxiv.org/abs/2003.13865

Yang, Z., Soltanian-Zadeh, S., & Farsiu, S. (2022). BiconNet: An edge-preserved connectivity-based approach for salient object detection. *Pattern Recognition*, *121*. https://doi.org/10.1016/j.patcog.2021.108231

Yu, C., & Helwig, E. J. (2022). The role of AI technology in prediction, diagnosis and treatment of colorectal cancer. *Artificial Intelligence Review*, *55*(1), 323–343. https://doi.org/10.1007/s10462-021-10034-y

Zhang, L., & Wen, Y. (2021). A transformer-based framework for automatic COVID19 diagnosis in chest CTs. *Proceedings of the IEEE International Conference on Computer Vision*, *2021-Octob*, 513–518. https://doi.org/10.1109/ICCVW54120.2021.00063

Zhang, Y., Gao, J., & Zhou, H. (2020). Breeds Classification with Deep Convolutional Neural Network. *ACM International Conference Proceeding Series*, 145–151. https://doi.org/10.1145/3383972.3383975

Zhao, C., Shuai, R., Ma, L., Liu, W., & Wu, M. (2022). Improving cervical cancer classification with imbalanced datasets combining taming transformers with T2T-ViT. In *Multimedia Tools and Applications* (Issue 2). https://doi.org/10.1007/s11042-022-12670-0

Zhao, J., Xiao, X., Li, D., Chong, J., Kassam, Z., Chen, B., & Li, S. (2021). mfTrans-Net: Quantitative Measurement of Hepatocellular Carcinoma via Multi-Function Transformer Regression Network. *Lecture Notes in Computer Science (Including Subseries Lecture Notes in Artificial Intelligence and Lecture Notes in Bioinformatics)*, *12905 LNCS*(September), 75–84. https://doi.org/10.1007/978-3-030-87240-3_8

Zheng, S., Lu, J., Zhao, H., Zhu, X., Luo, Z., Wang, Y., Fu, Y., Feng, J., Xiang, T., Torr, P. H. S., & Zhang, L. (2021). Rethinking Semantic Segmentation from a Sequence-to-Sequence Perspective with Transformers. *Proceedings of the IEEE Computer Society Conference on Computer Vision and Pattern Recognition*, 6877–6886. https://doi.org/10.1109/CVPR46437.2021.00681

Zheng, Y., Gindra, R. H., Green, E. J., Burks, E. J., Betke, M., Beane, J. E., & Kolachalama, V. B. (2022). A graph-transformer for whole slide image classification. *IEEE Transactions on Medical Imaging*. https://doi.org/10.1109/TMI.2022.3176598

Zhou, B., Schlemper, J., Dey, N., Salehi, S. S. M., Liu, C., Duncan, J. S., & Sofka, M. (2022). *DSFormer: A Dual-domain Self-supervised Transformer for Accelerated Multi-contrast MRI Reconstruction*. http://arxiv.org/abs/2201.10776

Zhou, H.-Y., Guo, J., Zhang, Y., Yu, L., Wang, L., & Yu, Y. (2021). *nnFormer: Interleaved Transformer for Volumetric Segmentation*. *XX*(Xx), 1–10. http://arxiv.org/abs/2109.03201

Zhou, L., Zhou, Y., Corso, J. J., Socher, R., & Xiong, C. (2018). End-to-End Dense Video Captioning with Masked Transformer. *Proceedings of the IEEE Computer Society Conference on Computer Vision and Pattern Recognition*, 8739–8748. https://doi.org/10.1109/CVPR.2018.00911

Zhuangzhuang Zhang, W. Z. (2021). *Pyramid Medical Transformer for Medical Image Segmentation*.

*Corresponding Author*: Tel; +234-811-940-0230
Email Address: emerald.henry@stu.cu.edu.ng